\documentclass[review]{elsarticle}
\usepackage{lineno,hyperref}
\modulolinenumbers[5]

\journal{Journal of Pattern Recognition}
\usepackage[numbers]{natbib}
\usepackage{framed, multirow, array}
\usepackage{graphicx}
\usepackage{threeparttable}
\usepackage{color,soul}
\usepackage{caption}
\usepackage{amssymb}
\usepackage{amsmath}
\usepackage{latexsym}
\usepackage{pifont}
\usepackage{tabularx,arydshln}
\usepackage{url}
\usepackage{xcolor}
\usepackage{subfigure}
\usepackage{multicol}
\usepackage{wrapfig}
\definecolor{newcolor}{rgb}{.8,.349,.1}
\newcommand{\xmark}{\ding{55}}%
\newcommand{\cmark}{\ding{51}}%

\newcommand{\bC}{\mathbf{C}}
\newcommand{\bD}{\mathbf{D}}
\newcommand{\bK}{\mathbf{K}}

\newcommand{\bbR}{\mathbb{R}}
\newcommand{\bb}{\mathbf{b}}
\newcommand{\bv}{\mathbf{v}}
\newcommand{\bx}{\mathbf{x}}
\newcommand{\by}{\mathbf{y}}

\newcommand{\bP}{\mathbf{P}}
\newcommand{\bR}{\mathbf{R}}

\newcommand{\bt}{\mathbf{t}}

\newcommand{\btx}{\mathbf{\tilde{x}}}
\newcommand{\btD}{\mathbf{\tilde{D}}}
\newcommand{\td}{{\tilde{d}}}

\usepackage{environ}         
\usepackage{etoolbox}        
\usepackage{graphicx}        

\newlength{\myl}
\let\origequation=\equation
\let\origendequation=\endequation

\RenewEnviron{equation}{
  \settowidth{\myl}{$\BODY$}                       
  \origequation
  \ifdimcomp{\the\linewidth}{>}{\the\myl}
  {\ensuremath{\BODY}}                             
  {\resizebox{\linewidth}{!}{\ensuremath{\BODY}}}  
  \origendequation
}

\begin{document}

\pagenumbering{gobble}
\clearpage

\let\WriteBookmarks\relax
\def\floatpagepagefraction{1}
\def\textpagefraction{.001}
\clearpage
\pagenumbering{arabic}

\title{Depth Perspective-aware Multiple Object Tracking}

\author[1]{Kha Gia {Quach}} 

\author[2]{Huu {Le}}

\author[3]{Pha {Nguyen}}

\author[1]{Chi Nhan {Duong}}

\author[1]{Tien Dai {Bui}}

\author[3]{Khoa {Luu} }

\address[1]{Computer Science and Software Engineering Department, Concordia University, Montreal, QC, CANADA}
\address[2]{Department of Electrical Engineering, Chalmers University of Technology, Sweden}
\address[3]{Computer Science and Computer Engineering, University of Arkansas, USA}

\begin{abstract}
This paper aims to tackle Multiple Object Tracking (MOT), an important problem in computer vision but remains challenging due to many practical issues, especially occlusions.
Indeed, we propose a new real-time Depth Perspective-aware Multiple Object Tracking (DP-MOT) approach\footnote{The implementation of DP-MOT will be publicly available.} to tackle the occlusion problem  in MOT. A simple yet efficient Subject-Ordered Depth Estimation (SODE) is first proposed to automatically order the depth positions of detected subjects in a 2D scene in an unsupervised manner. 
Using the output from SODE, a new Active pseudo-3D Kalman filter, a simple but effective extension of Kalman filter with dynamic control variables, is then proposed to dynamically update the movement of objects. In addition, a new high-order association approach is presented in the data association step to incorporate first-order and second-order relationships between the detected objects. The proposed approach consistently achieves state-of-the-art performance compared to recent MOT methods on standard MOT benchmarks.
\end{abstract}

\maketitle

\section{Introduction}

Multiple Object Tracking (MOT) 
has long been considered as one of the most important problems in computer vision, which has recently received much attention due to its potential in many relevant real-world applications.
In practice, solving MOT efficiently and robustly is indeed a challenging task due to missing or inaccurately detected objects, occlusions, and overlapping in crowded scenes. 
Among those factors, occlusions caused by object movements or detectors' failure have been shown to greatly deteriorate the overall performance of existing MOT solvers \cite{kim2015multiple}. Our work tackles some crucial drawbacks in current MOT solvers, especially occlusions, by proposing a simple yet effective geometric-based approach that
leverages perspective information from the scene.

To track multiple targets, e.g. pedestrians on the move, a common paradigm is track-by-detection, which primarily has two main steps: (1) detecting objects in each frame (2) assigning ID for each detection by linking them across multiple frames. 
Our work focuses on the second stage, \textit{i.e. data association and tracking}. Hence, our algorithm can be used on top of any existing object detectors.
Most MOT methods \cite{ren2018, Chen_2018_ECCV, maksai2019, sun2019deep, xu2019deepmot, wang2019towards, bergmann2019tracking, Li_2020_WACV, chu2019famnet,zhu2018online} that solely rely on the detection results, i.e. 2D bounding boxes and their associated features, 
could lead to ID switching between tracked objects, 
when an object is occluded in the scene, either by the other objects or scene backgrounds. 
This challenging occlusion problem can be technically handled with the support of perspective or depth information from the scene. 
Hence, by tracking 3D real-world coordinates of the objects, 3DMOT methods \cite{weng2019baseline, hu2019joint, scheidegger2018mono} can provide better results. 
Unfortunately, the use of such 3D information usually comes with an expensive cost. In particular, if depth sensors are not available, 
a monocular depth estimation method is needed. Thus, during the training stage, such methods often require both video frames and depth information whose ground truth labels are sometimes unavailable or expensive in practice. Moreover, their 3D-based algorithms are also computationally expensive that could also affect the real-time performance in many algorithms.

\begin{figure*}[t]
    \centering
    \includegraphics[width=1\columnwidth]{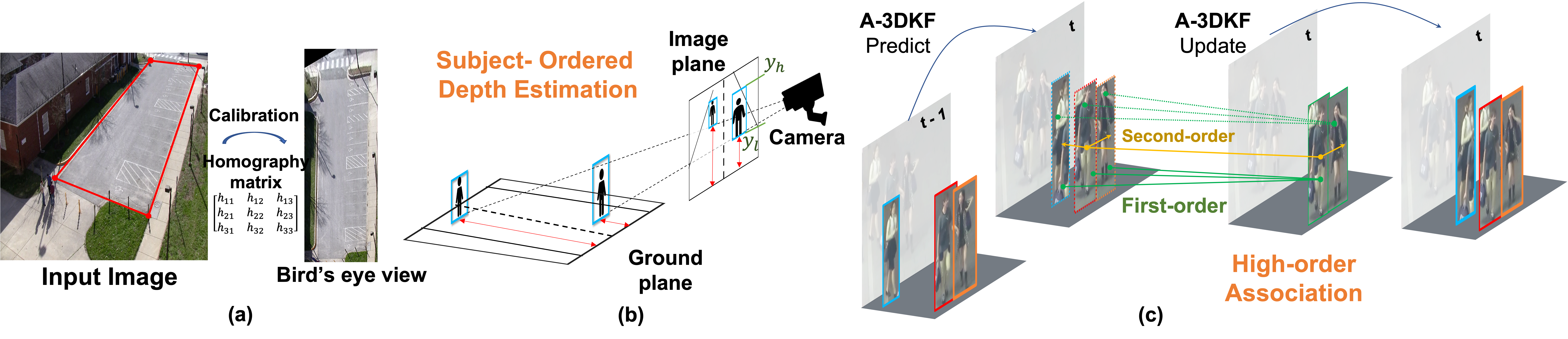}
    \caption{Illustrations of (a) Traditional calibration method (b) Subject-Ordered Depth Estimation and (c) High-order Association.}
    \label{fig:depth_estimation}
\end{figure*}{}

\textbf{Contributions of this Work:}
This paper introduces Depth Perspective-aware Multiple Object Tracking (DP-MOT) to solve the occlusion problems in MOT. Our contributions are four-fold.
\begin{itemize}
    \item A simple yet efficient \textit{Subject-Ordered Depth Estimation} (SODE) method is derived from a camera model and geometry-based technique 
    to approximate the depth positions of the detected subjects provided by any detector. 
    \item
    A new \textit{Active pseudo-3D Kalman filter} method with an \textit{adaptive control variable} provides dynamic and stable tracking of object movements w.r.t depth ordering. 
    \item
    A new \textit{high-order association} matrix is computed by measuring \textit{first-order} and \textit{second-order} similarity between the detected objects and tracks to improve the association step in handling occlusions (as shown in Fig. \ref{fig:depth_estimation} (c)).
    \item
    The proposed DP-MOT method consistently \textit{achieves the state-of-the-art} (SOTA) results compared to recent MOT approaches \cite{bergmann2019tracking, chu2019famnet, chu2019online} on the standard MOT benchmarks.
\end{itemize}

\textbf{SODE v.s. Traditional Calibration Methods:} Although the presented SODE is simple and performs in an unsupervised manner, it can achieve high accuracy in real-time. More important, it is quite different from traditional methods (Fig. \ref{fig:depth_estimation}(a)) where \textit{manual calibration} is required to transform to bird's eye view. Indeed, SODE can generally work for any new videos \textit{without calibration} while the traditional methods require a manual calibration in every new setup between two views.

\textbf{Assumptions in SODE:}  
Given a set of 2D frames in a video captured with a \textit{fixed camera height and rotation}, this work aims to tackle the occlusion problem in real-time using a new simple geometry-based depth perspective-aware approach (see Fig. \ref{fig:depth_estimation} (b)).
Particularly, far apart from prior work, the proposed method \textit{does not require} depth or 3D boxes as ground truth. It can be used as an ``add-on" component to support other tracking methods with any detector to improve their performance.

\textbf{Human Detection and Appearance Extraction in SODE:} It is noticed that SODE can simply work with any off-the-shelf deep learning methods for human detection and subject appearance deep feature extraction. The development of human detection and deep feature extraction is not in the scope of this work. In the experiments in this work, deep Feature Embedding Network (FEN) derived from \cite{wang2019towards} is used to extract appearance features for the detected boxes to obtain enriched objects.

\section{Related Works}

This section reviews some recent works on track-by-detect MOT algorithms which are summarized in Table~\ref{tb:MethodReview}. 
Generally speaking, the algorithms can be categorized as follows:

\textbf{Tracking by solving the matching problem:}
The Hungarian algorithm has been used in several methods as a matcher to associate the detection by some kinds of appearance features. 
Kim et al. \cite{kim2015multiple} proposed to use colors as the appearance features.
Some other methods try to improve the accuracy of the matcher by anticipating the locations of objects in the next frame. Ren et al. \cite{ren2018} and Chen et al. \cite{Chen_2018_ECCV} introduced reinforcement learning to predict the movement of objects. 
Furthermore, Maksai et al. \cite{maksai2019} applied a bi-directional Long Short-Term Memory (LSTM) to estimate the motion of trajectories.
Quach et al. \cite{quach2021dyglip} proposed a dynamic graph to transform pre-computed Re-ID features to new context-aware ID features, hence it performs better matching and yields more accurate results. Similarly, Nguyen et al. \cite{Nguyen_2022_CVPR} proposed a novel global association graph model with Link Prediction approach to predict existing tracklets location and link detections with tracklets via cross-attention motion modeling and appearance re-identification. 
Recently, Li et al. \cite{LI2022108738} proposed an end-to-end identity association network that consists of a geometry refinement network and an identity verification module to perform data association and reason the mapping between the detections and tracklets.

\textbf{Tracking as an end-to-end framework:} 
Sun et al. \cite{sun2019deep} and Xu et al. \cite{xu2019deepmot} presented an assignment algorithm that allows the tracker to be optimized in an end-to-end framework. 
Wang et al. \cite{wang2019towards} introduced a shared weight model between the detection and the association stage with an improved triplet loss to push away embedding of different identities in one batch. 
Bergmann et al. \cite{bergmann2019tracking} (Tracktor) shows a simple approach by exploiting the bounding box regression of the object detector to guess the position of objects in the next frame in a high-frame-rate video sequence without camera motion. 
Chu and Ling \cite{chu2019famnet} designed a model having an affinity estimation and multi-dimensional assignment component for visual object tracking. Zhou et al. \cite{zhou2020tracking} employed a straightforward approach, which trained a network to predict the movement offset from the previous frame and then matched with the nearest tracklet center point.
Joint Detection Embedding (JDE) \cite{wang2019towards} is a fast, simple, yet effective approach, which solves detection and embedding representation tasks in a single network structure and a fused loss function.
Li and Gao et al. \cite{Li_2020_WACV} introduced a graph network for extracting appearance and motion similarity separately. Sun et al. \cite{sun2020simultaneous} constructed three networks to compute three matrices, representing the object's motion, object's type, and object's visibility, respectively for every matching step. Peng et al. \cite{peng2020chained} passed a couple of frames to a single network, which can generate the target's bounding box pair between two frames, therefore no association step is needed. Xu et al. \cite{xu2020train} (DeepMOT) formulated MOT metrics as a loss function to construct a full trainable tracking procedure.
Yin et al. \cite{yin2020unified} trained a Siamese neural network for the joint task of single object tracking and multiple object association simultaneously.
Chan et al. \cite{CHAN2022108793} proposed an end-to-end network for detecting and tracking multiple object simultaneously.

\textbf{Tracking with 3D information} Some tracking methods using additional data from multiple cameras or special GPS or IMU sensors to solve the occlusion problem. Hu et al. \cite{hu2019joint} relied on ego-motion provided by a sensor to predict a vehicle's future location and used a weighted bipartite matching algorithm on affinities between tracks and detections. 
Weng et al. \cite{weng2019baseline} use LiDAR point cloud to obtain oriented 3D boxes from 3D object detector, then use a basic 3D Kalman filter to estimate objects' state and the Hungarian algorithm to associate tracks with new detection.
Scheidegger et al. \cite{scheidegger2018mono} exploited the 3D information of surrounding pedestrians to achieve high accurate predictions. Weng et al. \cite{weng2020gnn3dmot} (GNN3DMOT) took full advantage of both 2D and 3D motion and appearance features, then constructed a graph neural network to compute the affinity matrix.

These physics-based approaches, however, are also prone to failure with sensor noises and uncertainty in the detection of occluded people which often occur in real-world scenarios.

\begin{table*}[t]
    \small
	\footnotesize
	\centering
	\caption{The comparison of the properties between our online MOT approach and others.
	Kalman Filter (KF), Requirement (Req.), Dynamic (Dyn.), Movement (Mov.), Update (Upd.), first-order (FO) and second-order (SO), Real-time (RT), Yes (Y), No (N). 
	``$-$'' denotes an unknown property. 
	} 
	\resizebox{1.0\textwidth}{!}{
	\begin{tabular}{l| c| c c| c c| c}
		\textbf{Methods}  &
		\textbf{Domain}& \textbf{Depth Aware} & \textbf{3D Label Req.}& \textbf{Association}& \textbf{Dyn. Mov. Upd.}& \begin{tabular}{@{}c@{}}\textbf{RT}\end{tabular}\\
		\hline
		\hline
		Tracktor \cite{bergmann2019tracking} & \multirow{7}{*}{2D} & \xmark & N & FO & \xmark & \xmark \\
		CenterTrack \cite{zhou2020tracking} & & \xmark & N & FO & \xmark & \cmark \\
		JDE \cite{wang2019towards} &  & \xmark & N & FO & \xmark & \cmark \\
		DeepMOT \cite{xu2019deepmot} &  & \xmark & N & FO & \xmark & \xmark \\
		FAMNet \cite{chu2019famnet} &  & \xmark & N & PW  & \xmark  &  \xmark \\
		STRN \cite{xu2019spatial} &  & \xmark & N & Graph & \xmark  & $-$ \\
		MOTDT \cite{chen2018real} & & \xmark & N & FO & \xmark & \xmark \\
		\hline
		AB3DMOT \cite{weng2019baseline} & \multirow{3}{*}{3D} & \cmark & Y &  FO & \xmark & \cmark\\
		3DT \cite{hu2019joint} &  & \cmark & Y & FO & \cmark (LSTM) & \xmark  \\
		3D-PMBM \cite{scheidegger2018mono} &  & \cmark & Y & $-$& \xmark & \cmark\\
		\hline
		\hline
		\textbf{Ours} & \textbf{2D} & \textbf{\cmark} & \textbf{N} & \begin{tabular}{@{}c@{}}\textbf{High-order}\\ \textbf{(FO+SO)} \end{tabular}& \begin{tabular}{@{}c@{}}\cmark\\\textbf{(A-3DKF)}\end{tabular} & \cmark \\
	\end{tabular}
	}
	\vspace{-5mm}
		\label{tb:MethodReview} 
\end{table*}

\section{The Proposed DP-MOT Approach}

\begin{figure*} [!t]
    \centering
    \includegraphics[width=1\columnwidth]{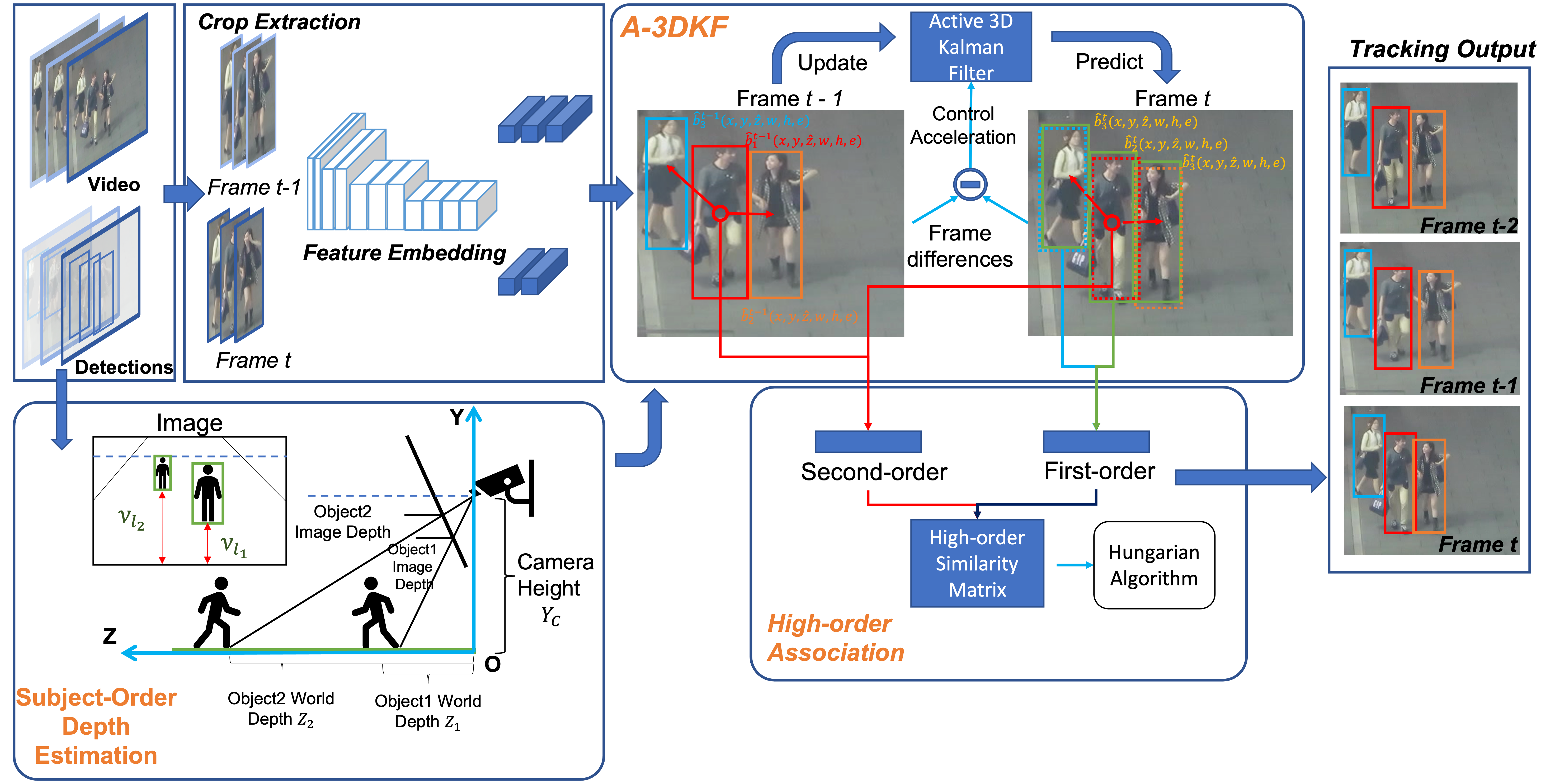}
    \caption{Our proposed framework with three main components: the subject-order depth estimation, the Active pseudo-3D Kalman filter, and the high-order association. The subject-order depth estimation approximates the ordering of objects in the scene according to their distance from the camera using only provided bounding box from the detector. In the active pseudo-3D Kalman filter, solid color boxes, i.e. blue, red and orange, represent track ids while dash color boxes represent the predicted location of tracked objects in the current frame $t$. Green color boxes are detected objects in the current frame $t$.}
    \label{fig:overall_architecture}
\end{figure*}

This section will detail three main components in the proposed framework as shown in Fig. \ref{fig:overall_architecture}, including the Subject-Ordered Depth Estimation to estimate depth ordering of bounding boxes, the Active pseudo-3D Kalman Filter component for tracking object positions, and the High-order Association for computing first-order and second-order perspective-aware association cost matrices.

\subsection{Problem Definition}
\label{sec:problem_def}

Assume that the video sequence contains $N$ frames, and there are $K$ objects that need to be tracked, we define the trajectory of the $k$-th object as $ \small T_k  = \{ \textbf{b}_k^{0}, \textbf{b}_k^{1}, \cdots, \textbf{b}_k^{t}, \cdots, \textbf{b}_k^{N} \}$, and $ \small \textbf{b}_k^t = (x, y, w, h)$, where $t$ denotes the frame number, the pair $(x, y)$ specifies the top-left location of the bounding box, 
while $w$ and $h$ respectively are the width and height of the detected object. 
Let $\mathcal{B}^{t} = \{\textbf{b}_1^{t}, \textbf{b}_2^{t}, \cdots, \textbf{b}_k^{t}, \cdots, \textbf{b}_{N_{\mathcal{O}}}^{t} \}$ denotes a set of $N_{\mathcal{O}}$ detected objects at the $t$-th frame.
The pipeline in this work consists of three main steps:

Firstly, the depth estimation SODE module (Section \ref{ssec:geo_depth}) 
provides the estimated depth $\hat{z}$ based on 2D geometry information of the boxes. 
Then, FEN \cite{wang2019towards} is used to extract appearance features $\mathbf{e}$ $\in \bbR^{512 \times 1}$ to obtain objects \\ $\hat{\bb}_k^{t} = \{ x, y, \hat{z}, w, h, \mathbf{e} \}$.
Secondly, the Active pseudo-3D Kalman filter (Section \ref{ssec:3d_kalman_filter}) will help to keep track trajectories $T_k$ up-to-date and predict new 3D-location of the tracked objects in the next frame. 
Finally, 
we compute a cost matrix  $\bC \in \bbR^{N_{\mathcal{T}} \times N_{\mathcal{O}}}$ ($N_{\mathcal{T}}$ and $N_{\mathcal{O}}$ are the number of tracked objects in frame $t-1$ and the number of detected objects in frame  $t$, respectively), by considering appearance similarity and 3D-spatial distance between each pair of predicted track $T_k$ from frame $t-1$ and detected object $\hat{\bb}_i^{t}$ in frame $t$. 
The computing of this matrix $\bC$ details in Section \ref{ssec:high-order_association}.
Based on this cost matrix $\bC$, the association problem is solved using the Hungarian algorithm to either properly assign detected objects to their corresponding existing tracks $T_k$, or to create new tracks for new objects. We also remove lost tracks when objects exit the scene as such tracks cannot match with any detected objects for several frames. 

\subsection{Subject-ordered Depth Estimation}
\label{ssec:geo_depth}

Given the shortcomings of existing learning-based monocular depth estimation approaches, we propose a novel mechanism to \textit{mathematically} approximate depth ordering, which can be used on top of any existing MOT trackers. Motivated by the common camera setup in most tracking applications such as video surveillance (see Fig.~\ref{fig:depth_estimation} (b)), we revisit a classical geometry-based technique to directly estimate depth ordering of the detected bounding boxes.

To utilize the setup as shown in Fig.~\ref{fig:overall_architecture}, we choose the world coordinate of 3D world points $(x, y, z)$ defined by $\mathbf{X}, \mathbf{Y}, \mathbf{Z}$ axes and origin $\mathbf{O}$ such that the ground plane (the plane where objects reside) coincides with the plane $y = 0$, i.e., an object's height is represented by its $y$ coordinate, while $z$ represents its depth, i.e. distance from the origin (see Fig.~\ref{fig:overall_architecture}). 
This simple setup for the world coordinate allows the camera pose to be represented by only camera height $h$ and pitch angle $\theta_X$ to form a camera coordinate frame with $\mathbf{X_c}, \mathbf{Y_c}, \mathbf{Z_c}$ axes, and origin $\mathbf{C}$. 

Let $\bK \in \bbR^{3\times 3}$ and $\bP = [\bR | \bt] \in \bbR^{3\times 4}$ denote the camera intrinsic and extrinsic matrices, respectively. 
Given the setup in Fig.~\ref{fig:depth_estimation} (b), the rotation matrix $\bR$ contains components along the $\mathbf{X}, \mathbf{Y}$ and $\mathbf{Z}$ axes as we allow small errors for the roll $\theta_Z$ and yaw $\theta_Y$ angles, while the translation vector $\bt = \left[0, \mathbf{Y_c}, 0\right]$. 

The projection of a point $(x, y, z, 1)$ in the homogeneous coordinates to the image coordinate $(u, v, 1)$ then reads:
\begin{eqnarray}
    \label{eq:camear_projection}
    \small
    & \begin{bmatrix}
    u \\ v \\ 1
    \end{bmatrix} = \frac{1}{Z} 
    \bK \bP \begin{bmatrix}
    x \\ y \\ z \\ 1
    \end{bmatrix} \\
    & = \frac{1}{Z} \begin{bmatrix}
    f & 0 & u_c \\
    0 &  f & v_c \\
    0 & 0 & 1 \\
    \end{bmatrix} \times 
    \begin{bmatrix}
    r_1 && r_2 && r_3 && 0 \\
    r_4 && r_5 && r_6 && \mathbf{Y_c} \\
    r_7 && r_8 && r_9 && 0 \\
    \end{bmatrix} \times
    \begin{bmatrix}
    x \\ y \\ z \\ 1
    \end{bmatrix} \nonumber
\end{eqnarray}
where the rotation matrix is defined as

\resizebox{.8\linewidth}{!}{
\begin{minipage}{\linewidth}
\begin{eqnarray}
    \footnotesize
    &  \bR = \bR_z(\theta_Z) \bR_y(\theta_Y) \bR_x(\theta_X) =  
    \begin{bmatrix}
    r_1 && r_2 && r_3 \\
    r_4 && r_5 && r_6 \\
    r_7 && r_8 && r_9 \\
    \end{bmatrix} \\
    & = \begin{bmatrix}
    \cos \theta_Z && - \sin \theta_Z && 0 \\
    \sin \theta_Z && \cos \theta_Z && 0 \\
    0 && 0 && 1 \\
    \end{bmatrix} \times
    \begin{bmatrix}
    \cos \theta_Y && 0 && \sin \theta_Y \\
    0 && 1 && 0 \\
    \sin \theta_Y && 0 && \cos \theta_Y \\
    \end{bmatrix} \times
    \begin{bmatrix}
    1 && 0 && 0 \\
    0 && \cos \theta_X && -\sin \theta_X \\
    0 && \sin \theta_X && \cos \theta_X \\
    \end{bmatrix} \nonumber \\
    & =
    \begin{bmatrix}
    \cos \theta_Z \cos \theta_Y && \cos \theta_Z \sin \theta_Y \sin \theta_X - \sin \theta_Z \cos \theta_X && \cos \theta_Z \sin \theta_Y \cos \theta_X + \sin \theta_Z \sin \theta_X  \\
    \sin \theta_Z \cos \theta_Y && \sin \theta_Z \sin \theta_Y \sin \theta_X + \cos \theta_Z \cos \theta_X && \sin \theta_Z \sin \theta_Y \cos \theta_X - \cos \theta_Z \sin \theta_X \\
    -\sin \theta_Y && \cos \theta_Y \sin \theta_X && \cos \theta_Y \cos \theta_X \nonumber  \\
    \end{bmatrix} 
\end{eqnarray}
\end{minipage}
}

Expanding the terms in~\eqref{eq:camear_projection} for the $v$ coordinate, and using some basic algebraic manipulations, we obtain the relation between the object's position $y$, $z$ in the world coordinate and the image coordinate $v$ as follows (note that we write $y$ as a function of $v$):

\begin{equation}
    \small
    y (v) = \frac{z \left( (v - v_c) r_9 - fr_6 \right) - f \mathbf{Y_c} + x \left( (v - v_c) r_7 - f r_4 \right) }
    {(v_c - v) r_8 + f r_5}.
    \label{eq:Y_relation}
\end{equation}
Recall that, in most tracking applications, it is common to track objects moving on the grounds. The grounds can be either flat, slope, or bumpy, however, to simplify the problem, we use a plane to approximate the grounds. Thus, the object's translation in the $\mathbf{Y}$-axis is approximately equal to $0$.
Under this approximation, the bottom of the detected bounding boxes corresponds to the coordinates $y=0$ in the real-world. Thus, we have $y(v_l) = 0$,
where $v_l$ represents the $v$ coordinate at the bottom of a particular detected bounding box. 
The estimated object depth can be then derived as in Eqn. (\ref{eq:Z_estimate}).
\begin{equation}
    \small
    \bar{z} = \frac{f \mathbf{Y}_c +  x \left(f r_4 + (v_c - v_l) r_7  \right) }{ (v_l - v_c) r_9 - f r_6 }.
    \label{eq:Z_estimate}
\end{equation}
We choose the world coordinate of 3D world points such that $\theta_Y$ and $\theta_Z$ equal to 0. Thus, $r_4$ and $r_7$ equal to 0 and we can omit the second term in Eq. \ref{eq:Z_estimate}.
This shows that the approximated depth $\bar{z}$ of a 3D point $(x, y, z)$ and its corresponding image coordinate $v_l$ appear to be co-linear as we consider other values are constant and relatively small.
As a result, we can directly use $v_l$, i.e. the bottom line of the objects' bounding boxes, to estimate the depth ordering between objects. 
Our idea here is to relax the need of estimate absolution depth value $z$ in meters. In other words, our estimated depth order $\hat{z}$ of an object box $\bb_k^t$ is defined as a quantized depth value $z$ as $\hat{z} = q( \frac{1}{v_l})$. 
Setting up the world coordinate to have zero yaw and roll angles in $\bP$ should be relatively easy via standard camera calibration in practice, compared to the huge amount of labeled training data required by other learning-based methods.
In the case of moving cameras, e.g. handheld cameras or cameras mounted on vehicles, the world coordinate is replaced by an ego-coordinate system with the origin $\mathbf{O}$ at the center bottom of the moving car/person holding the camera at the current time $t$. As a result, we can rewrite extrinsic matrix $\bP^t$ as
\begin{eqnarray}
    \label{eq:camear_projection}
    \small
    & \bP^t  = 
    \begin{bmatrix}
    1 && 0 && 0 && 0 \\
    0 && 1 && 0 && \mathbf{Y_c} \\
    0 && 0 && 1 && 0 \\
    \end{bmatrix}
\end{eqnarray}
Since we only consider objects' locations at time $t$ and $t+1$ while performing online tracking, we can approximate those coordinates with the same extrinsic matrix, i.e. $\bP^t \approx \bP^{t+1}$.
Then, the estimated object depth can be rewritten as in Eqn. (\ref{eq:Z_estimate_new}).
\begin{equation}
    \small
    \bar{z} = \frac{f \mathbf{Y}_c}{ (v_l - v_c) }.
    \label{eq:Z_estimate_new}
\end{equation}
Thus, we can also obtain estimated depth order $\hat{z}$ of an object box $\bb_k^t$ as $\hat{z} = q( \frac{1}{v_l})$. Therefore, our depth ordering estimation is applicable in the case of dynamic cameras.
We conduct some experiments using this depth ordering computation. Surprisingly, it still works well in different scenarios, including both static and moving cameras, on sequences in both MOT challenge \cite{MOTChallenge2015, MOT16} and nuScene dataset \cite{caesar2020nuscenes}.

In the next Section \ref{ssec:3d_kalman_filter}, we will show how the estimated depth ordering information can be integrated into our active pseudo-3D Kalman filter to significantly improve tracking performance. 
Furthermore, the pseudo-3D Kalman filter can help to correct missed detected objects due to human jumping or inaccurately detected boxes.

\subsection{Active pseudo-3D Kalman Filter}
\label{ssec:3d_kalman_filter}

Based on the refined objects with the estimated depth $\hat{z}$, we propose an extension of Kalman filter (KF), named Active pseudo-3D Kalman Filter (A-p3DKF), to significantly improve the tracking accuracy. In particular, each object in the scene is associated with a KF tracker, $\bK_t[i], i = 1 \dots N_{\mathcal{T}}$, that actively maintains and updates its state. 
In the following, we describe the configurations of a particular tracker.

Given depth information for each detected bounding box, we formulate states for each tracker as $\mathbf{s}_{t,t}[i] = [c_x, c_y, \hat{z}, s, r, c'_x, c'_y, \hat{z}', s']$, 
where $c_x, c_y$ specify the center of the box, $\hat{z}$ is the \emph{estimated} depth of the object computed based on~\eqref{eq:Z_estimate}, 
and $s, r$ are scale and ratio parameters, respectively. The remaining are the derivatives of box location and scale, i.e., the velocity of those measures. We use double subscript $ \mathbf{s}_{{t_1}, {t_2}}$ to denote that $ \mathbf{s} $ is predicted at time step $t_1$ and updated at $t_2$, respectively. 
Unlike traditional pseudo-3D Kalman filters, we proposed to add a control variable, i.e., acceleration, defined as $\hat{\mathbf{u}}_{t,t}$ = $[c''_x, c''_y, \hat{z}'']$  and a control matrix ($\mathbf{G}$).
Given an initial state $\mathbf{s}_{1, 0}$ and the co-variance (uncertainty) matrix $\mathbf{A}_{1, 0}$, we can first predict the next state $\mathbf{s}_{t+1, t}$ and uncertainty $\mathbf{A}_{t+1, t}$ at time $t+1$ taking into account sudden changes or movements as follows,
\begin{eqnarray}
        \small
        \mathbf{s}_{t+1, t} = \mathbf{F} \mathbf{s}_{t, t} + \mathbf{G} \hat{u}_{t,t},  \mathbf{A}_{t+1, t} = \mathbf{F} \mathbf{A}_{t, t} \mathbf{F}^T + \mathbf{Q},
\end{eqnarray}
where $\mathbf{F}$ and $\mathbf{G}$ are the state transition and the control matrix, 
and $\mathbf{Q}$ is the process noise uncertainty. 
\begin{equation}
    \small
\mathbf{G} = 
\begin{bmatrix} 
\text{diag}_{3 \times 3} (0.5 \Delta t^2) \\
\text{diag}_{3 \times 3} (\Delta t) \\
\end{bmatrix}, 
\mathbf{Q} = \mathbf{G}\sigma^2\mathbf{G}^T,
\Delta t = 1.
\label{eq:G_define}
\end{equation}
We can compute the control variable $\hat{\mathbf{u}}_{t,t}$ from difference between three adjacent frames based on the optimized optical flow \cite{kroegerECCV2016}  
instead of pre-defined fixed acceleration values, e.g. set as zero for constant velocity. 
To reduce the computational cost of optical flow, we only compute flow between two small sets of key points, i.e. 16 from our experiments, limited within an extended region of detected objects.
This parameter $\hat{\mathbf{u}}_{t,t}$ can be estimated from the scene dynamics in the following steps.
\begin{itemize}
    \item \textit{Step 1}. Compute local optical flow of each tracker, around the location at ($c_x$, $c_y$), in three consecutive frames as ${\delta}^{t-1, t}$ and ${\delta}^{t, t+1}$.
    \item \textit{Step 2}. Compute control variable as \\ $\hat{\mathbf{u}}_{t,t} = [ {\delta}^{t, t+1}_x - {\delta}^{t-1, t}_x, {\delta}^{t, t+1}_y - {\delta}^{t-1, t}_y, \gamma ({\delta}^{t, t+1}_y - {\delta}^{t-1, t}_y) ]$, where $\gamma$ is the ratio obtained from Eq. \eqref{eq:Y_relation}.
\end{itemize}

Finally, we can correct the prediction of A-3DKF with a new predicted bounding box (measurement) at time $t$ in the following steps.
\begin{itemize}
    \item 
\textit{Step 1}. Compute the Kalman Gain \\ $ \mathbf{K}_t = \mathbf{A}_{t, t-1} \mathbf{H}^T (\mathbf{H} \mathbf{A}_{t, t-1} \mathbf{H}^T + \mathbf{R}_t )^{-1}, $
where $\mathbf{H}$ is the observation matrix and $\mathbf{R}_t$ is the measurement uncertainty. 
\item
\textit{Step 2}. Update state with measurement \\ $  \mathbf{s}_{t,t} = \mathbf{s}_{t, t-1} + \mathbf{K}_t ( \mathbf{z}_t - \mathbf{H} \mathbf{s}_{t, t-1} ) $, 
where $\mathbf{z}_t$ is the detector prediction.
\item
\textit{Step 3}. Update covariance matrix $\mathbf{A}_{t,t} = ( \mathbf{I} - \mathbf{K}_t\mathbf{H} ) \mathbf{A}_{t, t-1} $,
where $\mathbf{I}$ is the identity matrix.
\item
\textit{Step 4}. Predict the new state $\mathbf{s}_{t+1,t} =  \mathbf{F} \mathbf{s}_{t, t} + \mathbf{G} \hat{u}_{t,t} $,
\end{itemize}

\subsection{High-order Association Matrix Computing}
\label{ssec:high-order_association}

Given the enriched information obtained from our framework, i.e. the relative object depths, the velocity, and acceleration vectors predicted from the A-3DKF, this section describes a novel usage of such information to improve the association task. Unlike traditional similarity/cost matrix computation, our proposed cost matrix is computed based on two crucial pieces of information: \textit{first-order} and \textit{second-order} relationships between tracks and detections.
Our cost matrix $\bC \in \bbR^{N_{\mathcal{T}} \times N_{\mathcal{O}}}$ measures the similarity between $N_{\mathcal{T}}$ tracked objects predicted from A-3DKF from previous frame $t-1$ and $N_{\mathcal{O}}$ detected objects at current frame $t$.

\subsubsection{Matrix Computing based on First-order Relationship}
The first-order cost matrix $\mathbf{C}_{F} \in \bbR^{N_{\mathcal{T}} \times N_{\mathcal{O}}} $ is directly computed based on appearance features and depth-aware intersection over Union (d-IoU) distances between tracks and detections. These distances are formulated as two matrices, $\mathbf{C}_{a} \in \bbR^{N_{\mathcal{T}} \times N_{\mathcal{O}}}$ and $\mathbf{C}_{d-IoU} \in \bbR^{N_{\mathcal{T}} \times N_{\mathcal{O}}}$.
In the following, we describe the design of $\mathbf{C}_{a}$ and $\mathbf{C}_{d-IoU}$.

\paragraph{Appearance Matrix}
Our appearance matrix follows closely with recent works on multiple object tracking, where the score of associating the $i$-th track to the $j$-th detected object is assigned to the matrix element $\bC_a (i,j)$ and is computed by the cosine similarity between two appearance vectors, 
i.e., 
\begin{equation}
    \small
    \bC_a (i,j) = \frac{\mathbf{e}_i \cdot  \mathbf{e}_j}{ \| \mathbf{e}_i \| \cdot \| \mathbf{e}_j \| } 
\end{equation}
Note that here $\mathbf{e} \in \bbR^{512}$ is the appearance vector obtained from the FEN introduced in Section~\ref{sec:problem_def}.

\paragraph{Depth-aware Intersection over Union (IoU) Matrix}
Since we only estimate the depth ordering of the detected and tracked objects, we extend the traditional 2D overlapping cost matrix to IoU considered within an allowable range of distance in-depth, named d-IoU. The d-IoU matrix is defined as,
\begin{eqnarray}
    \small
    \bC_{d-IoU} (i, j) = \frac{\mathbf{b}_i \cap  \mathbf{b}_j}{ \mathbf{b}_i  \cup \mathbf{b}_j } \otimes \left( \|\hat{z}_i - \hat{z}_j \| < \tau_Z \right) 
\end{eqnarray}
where $\tau_Z$, which is experimentally set to $3$, is the ranging threshold for depth steps to consider objects having close proximity or overlapping.

\subsubsection{Matrix Computing based on Second-order Relationship}
The second-order cost matrix $\mathbf{C}_P \in \bbR^{N_{\mathcal{T}} \times N_{\mathcal{O}}} $ is computed based on appearance features and spatial features of each object/track with respect to other objects/tracks within a frame. 
This section describes our novel method to utilize the output of SODE and A-3DKF to compute \textit{second-order spatial and appearance distances}. 

Let us denote by $\bx_k = \{ c_x, c_y, \hat{z} \} \in \bbR^3$ and $\bv_k = \{ c'_x, c'_y, \hat{z}' \} \in \bbR^3$ the  location and the velocity vector of the $k$-th tracked object ($k = 1 \dots N_{\mathcal{T}}$) at time $t - 1$, respectively. 
Given $\bx_k$ and $\bv_k$ from our Active pseudo-3D Kalman filters, we can obtain the predicted location $\btx_k$ for the $k$-th tracked object. 
After obtaining the predicted 3D positions of the tracked objects, we denote by $\td_{ij}$ the distance between $\btx_i$ and $\btx_j$, i.e., $ \td_{ij} = \| \btx_i - \btx_j \| $.
We also define a vector $\btD_k = s([\td_{k1}, \dots, \td_{k(k-1)},d_{k(k+1)}, \dots \td_{k N_{\mathcal{T}}} ]^T)$ 
where the function $s(\cdot)$ sorts all the provided elements in ascending order. 
Intuitively, the vector $\btD_k$ associated with the $k$-th tracked object contains the sorted distances between itself to the remaining tracked objects.

Let $\{\by_{l}\}_{l=1}^{N_{\mathcal{O}}}$ denote the bounding boxes and estimated depth values returned by the detector and SODE at time step $t$ , where $\by_l = \{ c_x, c_y, \hat{z} \} \in \bbR^3$. Similarly, let $d_{pq} = \| \by_p - \by_q \|$ be the distance between the predicted bounding boxes $p$ and $q$, and ${\bD_l = s([d_{l1}, \dots, d_{l(l-1)},d_{l(l+1)}, \dots d_{lN_{\mathcal{O}}} ]^T) }$ be the sorted distances between the $l$-th detected bounding box to the remaining boxes. 
We assume that the number of objects and their assigned IDs remain unchanged from frame $t - 1$ to frame $t$ and the detected bounding boxes in frame $t$ match their predicted locations. Under such assumption, if tracked object $k$ matches the bounding box $l$, it is expected that $\|\btD_k - \bD_l\| = 0$. In this case, $\btD_k$ and $\bD_l$ act as useful feature vectors for the association task.

However, in practice, it usually happens that $N_{\mathcal{T}} \neq N_{\mathcal{O}}$ since new objects may appear in frame $t$, or some existing objects are no longer captured by the camera. In such scenarios, to obtain robust matching, algorithms such as graph matching have proven to be useful choices. However, employing such algorithms is computationally expensive, which may degrade the real-time performance of our tracking pipeline.
Therefore, we propose an approximation scheme to make use of the distance information while keeping the existing linear assignment framework. In particular, given two vectors $\btD \in \bbR^{N_{\mathcal{T}}}$ and $\bD \in \bbR^{N_{\mathcal{O}}}$ from a tracked object and a detected box, without loss of generality, let first consider $N_{\mathcal{T}} \le  N_{\mathcal{O}}$. We propose a new second-order function $f(\btD,\bD):\bbR^{N_{\mathcal{T}}} \times \bbR^{N_{\mathcal{O}}} \mapsto \bbR$ to measure the distance between $\btD$ and $\bD$ as follows,
\begin{eqnarray}
    \small
f(\btD,\bD) = \sum_{i=1}^{N_{\mathcal{T}}} \|\btD_{[i]} - \bD_{[j-1+i]}\|, \\ \text{where}\; j = \arg\min_{j=1 \dots N_{\mathcal{O}}}  \|\btD_{[1]} - \bD_{[j]}\| \nonumber
\end  {eqnarray}
where the notation $\bD_{[i]}$ denotes the $i$-th element of the vector $\bD$. Note that the above problem can be solved efficiently, since $\btD$ and $\bD$ are sorted vectors,  by first searching for $j$ using binary search in $O(\log N_{\mathcal{O}})$, and then compute the value of $f$ in $O(N_{\mathcal{T}})$.
In the case of $N_{\mathcal{T}} > N_{\mathcal{O}}$, we will ignore all the other $i$-th elements where $i > N_{\mathcal{O}}$.
In summary, our \textbf{\textit{second-order spatial distance matrix}} $\bC_{P_d}$ is defined as $\bC_{P_d} (i,j) = f(\btD_i, \bD_j)$.
In this way, we can incorporate a second-order spatial relationship between tracks and detections within a current frame and a constraint spatial relationship between weighted bipartite graph matching between current and previous frames.
Similarly, we can compute our \textit{\textbf{second-order appearance distance matrix}} $\bC_{P_a}$ as $ \bC_{P_a} (i,j) = f(\mathbf{\Tilde{A}}_i, \mathbf{A}_j). $
where $\mathbf{\Tilde{A}}_i$ and $\mathbf{A}_j$ are the sorted cosine distances between the appearance vector of tracked objects and detected bounding boxes to their remaining ones in frame $t$.

In summary, our cost matrix $\bC \in \bbR^{N_{\mathcal{T}} \times N_{\mathcal{O}}}$ is the combination of the following
\begin{equation}
    \small
    \label{eq:affinity_matrix}
    \mathbf{C} = \mathbf{C}_F+ \mathbf{C}_P = \alpha \left( \mathbf{C}_a + \mathbf{C}_{d-IoU} \right) + \beta \left(  \bC_{P_a} + \bC_{P_d} \right)
\end{equation}
The values of hyper-parameters $\alpha$ and $\beta$ which control the weights for each component, are manually chosen through experiments in the next section. 

\section{Experiment Results}

\subsection{Databases and Evaluation Metrics}

\paragraph{Datasets} To evaluate the effectiveness of the proposed DP-MOT, it is experimented on MOT Challenge \cite{MOTChallenge2015, MOT16} with the pedestrian tracking dataset. This dataset has two different versions, including MOT16 and MOT17. Each includes several real-world video sequences recorded on the streets with numerous occlusions and the challenging motion of people crossing and hiding each other. Videos have a diversity of views, illuminated conditions, and video frame rates. MOT16 and MOT17 share the same videos but have different annotation sets.
\textbf{nuScenes} \cite{caesar2020nuscenes} is one of the large-scale datasets for Autonomous Driving with 3D object annotations. It contains 1,000 videos of 20-second shots in a setup of 6 cameras, i.e. 3 front and 3 rear ones,  with a total of 1.4M images. It also provides 1.4M manually annotated 3D bounding boxes of 23 object classes based on LiDAR data. This dataset is an official split of 700, 150 and 150 videos for training, validation and testing, respectively.

\paragraph{Metrics} MOT methods are commonly evaluated in multiple criteria. Multiple Object Tracking Accuracy (MOTA) and ID F1 score (IDF1) are the most important MOT metrics together with Mostly Tracked objects (MT), Mostly Loss objects (ML), and ID switches (ID Sw.).

\begin{figure} [!t]
    \centering
    \includegraphics[width=0.6\columnwidth]{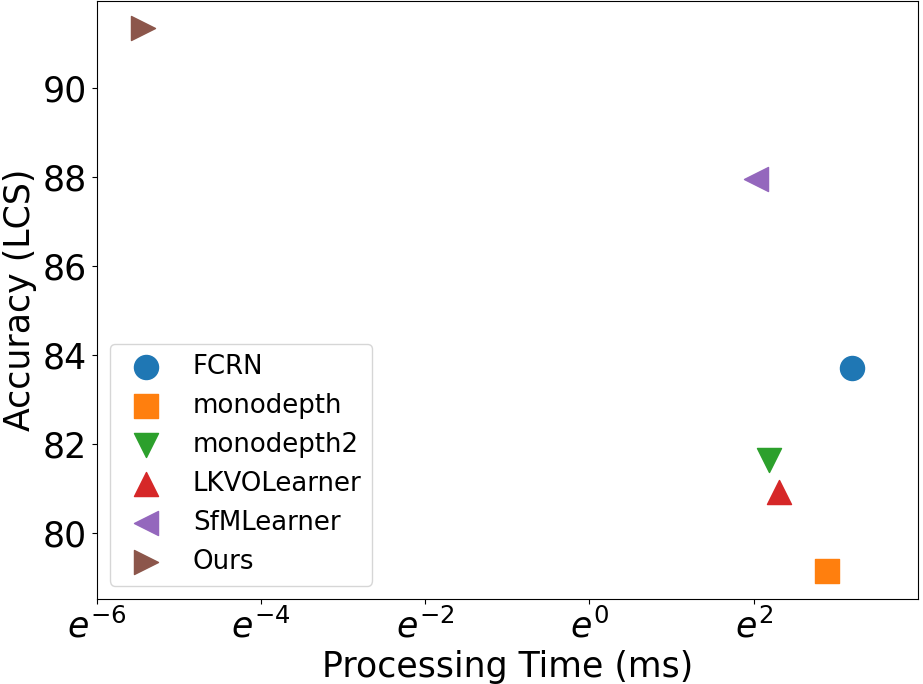}
    \caption{The comparisons of the depth estimation approaches, including FCRN \cite{laina2016deeper}, Monodepth \cite{godard2017unsupervised}, Monodepth2 \cite{godard2019digging}, LKVOLearner \cite{wang2018learning}, SfMLearner \cite{zhou2017unsupervised} and our proposed SODE on People RGB-D dataset. 
    The higher and more left one is better. \textbf{Best viewed in color.}}
    \label{fig:depth_val_comparision}
\end{figure}

\begin{figure*}
    \centering
    \includegraphics[width=1.0\columnwidth]{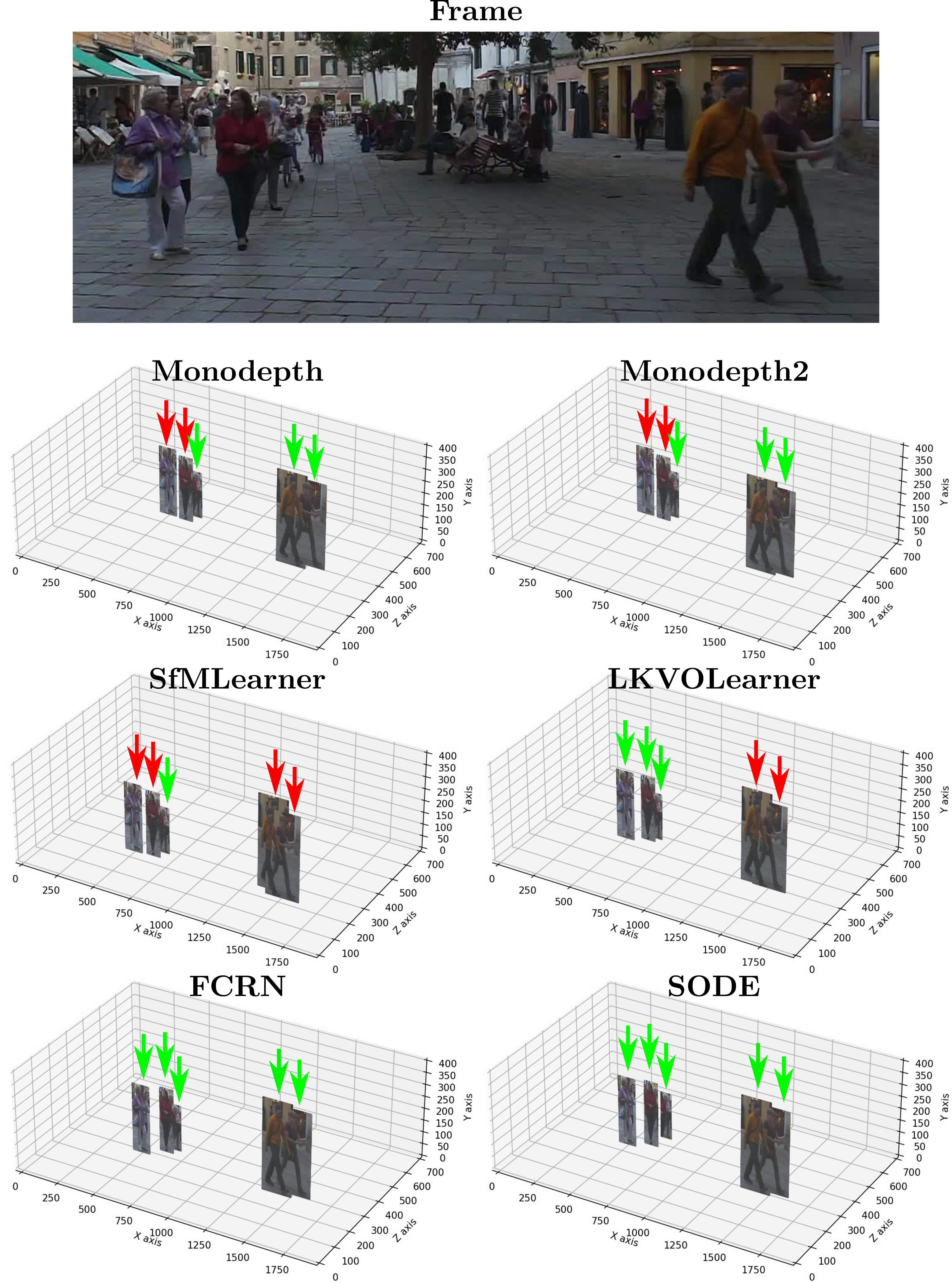}
    \caption{The visualization of the depth estimation approaches, including FCRN \cite{laina2016deeper}, Monodepth \cite{godard2017unsupervised}, Monodepth2 \cite{godard2019digging}, LKVOLearner \cite{wang2018learning}, SfMLearner \cite{zhou2017unsupervised} and our proposed SODE on MOT17 dataset. Green arrows indicate right positional correlations, and red arrows indicate wrong positional correlations on the $XZ$ plane. \textbf{Best viewed in color and zoom in.}}
    \label{fig:depth_vis_comparision}
\end{figure*}

\subsection{Ablation Studies}

In this section, we perform several ablation studies to show the role of each proposed module. 

\paragraph{\textbf{Comparisons of Subject Ordered Depth Estimation and SOTA Depth Estimation}}

This experiment aims to study the effectiveness of the proposed SODE method 
in estimating the orders of objects in the scene. 
Since we only need to approximate relative depth information, i.e. depth ordering between detected objects w.r.t. the camera, in this experiment we propose to evaluate the accuracy of our estimator by comparing the estimated relative depth to the ground-truth depths provided by the RGB-D People dataset \cite{spinello2011people, luber2011people}.

In particular, for each frame $t$, 
based on the ground-truth depth information, we sort the bounding boxes by their depths in the ascending order to obtain a set of indexes $d_{\text{truth}}$. With the same set of bounding boxes, we use our proposed method to estimate their relative depths, and sort the bounding boxes according to their estimated depths to obtain the vector $d_{\text{estimated}}$.
Then, the accuracy of depth orders is validated via the Longest Common Subsequence (LCS) metric between $d_{\text{estimated}}$ and $d_{\text{truth}}$ as $Accuraccy = \frac{LCS(d_{truth}, d_{estimated}) \times 100}{|d_{truth}|}$.
Fig. \ref{fig:depth_val_comparision} illustrates the achieved performance in terms of speed and accuracy of our SODE in comparison to other depth estimation methods on 2,000 randomly selected frames of the database. Since all methods also used the same detected bounding boxes in their estimation process, we exclude the detection time and only report the depth estimation step to obtain the subject-ordered list. As can be seen, our SODE method outperforms other approaches in computing the subject order w.r.t. depth property while maintaining real-time processing speed. The processing time of SODE step is fast due to its simplicity in computation. SODE method achieves relatively good accuracy thanks to re-using the learned "perspective" knowledge from the detector.
Fig. \ref{fig:depth_vis_comparision} presents the relative visualization of depth estimation methods. Notice that the outputs of other methods are the depth maps with exact depth values, so the depth value for each pedestrian is computed as the average depths of all pixels inside the bounding box of that subject and projected on the $XZ$ plane. We found that such a way of getting depth orders could have some negative impacts on many existing depth estimation approaches in our experiments due to overlapping with background and other foreground objects. In fact, our SODE is less likely to be confused by other background/foreground information as the only source of information is the "perspective" knowledge from the pre-trained pedestrian detector.

\begin{table*}[!t]
    \footnotesize
    \caption{Evaluation on MOT-17 (SDP detectors) validation set with provided boxes from SDP detector for different depth estimation approaches in terms of MOT metrics and depth ordering accuracy. $\downarrow$ means the smaller the better; $\uparrow$ means the larger the better.}
    \label{tab:ablation_depth_estimation}
    \centering
    \resizebox{\textwidth}{!}{
        \begin{tabular}{l|cc|ccc|c}
            \textbf{Depth models} & \textbf{MOTA} $\uparrow$ & \textbf{IDF1} $\uparrow$ & \textbf{MT} $\uparrow$ & \textbf{ML} $\downarrow$ & \textbf{ID Sw.} $\downarrow$ & \textbf{LCS Acc.} $\uparrow$ \\
             \hline
             \hline
             Monodepth \cite{godard2017unsupervised} & 63.9 & 68.2 & 142 & 152 & 975 & 79 \\
             Monodepth2 \cite{godard2019digging} & 64.1 & 70.2 & 145 & 154 & 862 & 81.5\\
             SfMLearner \cite{zhou2017unsupervised} & 64.7 & 70.7 & 151 & 155 & 537 & 88.2\\
             LKVOLearner \cite{wang2018learning} & 64.9 & 72.4 & 153 & 154 & 507 & 81.3\\
             FCRN \cite{laina2016deeper} & 65.5 & 72.2 & 157 & 145 & \textbf{330} & 84.1\\
             Our SODE & \textbf{66.6} & \textbf{72.8} & \textbf{179} &  \textbf{142} & 399 & \textbf{91.8} \\
        \end{tabular}
    }
\end{table*}

To further evaluate our proposed SODE, we conduct another experiment on MOT-17 validation set using different estimated depth values from other depth estimation approaches in our MOT framework. In such a way, we will demonstrate that by using the proposed depth estimation method, we can observe similar or better performance than directly using the absolute depth, i.e. in meters, information. The results are shown in Table \ref{tab:ablation_depth_estimation}. As we can see in the last column of Table \ref{tab:ablation_depth_estimation}, the depth orders provide by other depth estimation methods are less accurate than our SODE method. Thus, it reduces the performance of our MOT framework since we are using the same tracking approach as described in Sections \ref{ssec:3d_kalman_filter} and \ref{ssec:high-order_association}. Such a gap in the accuracy of depth orders is because of the depth orders estimated using the average depths of all pixels inside the bounding box of an object. The depth information within the bounding box can be redundant and noisy.

\begin{table*}[!t]
    \footnotesize
    \caption{Evaluation on MOT-16 (FRCNN detector) and MOT-17 (SDP detectors) validation set with provided boxes from various detectors for improvements with KF using 3D information, acceleration, and high-order similarity matrix. $\downarrow$ means the smaller the better; $\uparrow$ means the larger the better.}
    \label{tab:ablation_KF_Association}
    \centering
    \resizebox{\textwidth}{!}{
        \begin{tabular}{l|lc|cc|ccc}
            \textbf{Motion models} & \textbf{Association} & \textbf{Detector} & \textbf{MOTA} $\uparrow$ & \textbf{IDF1} $\uparrow$ & \textbf{MT} $\uparrow$ & \textbf{ML} $\downarrow$ & \textbf{ID Sw.} $\downarrow$ \\
             \hline
             \hline
             2DKF & First-order & FRCNN & 54.7 & 55.3 & 25.7 & 26.8 & 1254 \\
             2DKF & \textbf{High-order} & FRCNN & 57.7 & 57.3 & 30.6 & 23.6 & 1172 \\
             3DKF & First-order & FRCNN & 57.6 & 56.2 & 28.8 & 30.7 & 955 \\
             3DKF & \textbf{High-order} & FRCNN & 58.7 & 58.2 & 32.1 & 21.7 & 903 \\
             A-2DKF & First-order & FRCNN & 57 & 55.5 & 28.63 & 25.1 & 1524 \\
             A-2DKF & \textbf{High-order} & FRCNN & 57.2 & 57.6 & 29 & 24.95 & 819 \\
             \textbf{A-3DKF }& First-order & FRCNN &  58.6 & 58.3 & 29.8 & 22.4 & 862 \\
             \textbf{A-3DKF} & \textbf{High-order} & FRCNN &  \textbf{58.9} & \textbf{59.6} & \textbf{31.33} & \textbf{21.6} & \textbf{821} \\ 
             \hline
             2DKF & First-order & SDP & 68.6 & 63.7 & 39.4 & 16.7 & 1275 \\
             2DKF & \textbf{High-order} & SDP & 70.8 & 65.2 & 46.2 &  13.6 & 1023\\
             3DKF & First-order & SDP & 71.1 & 66.5 & 45.4 & 13.2 & 928 \\
             3DKF & \textbf{High-order} & SDP & 72.4 & 66.8 & 48.2 & 12.5 & 773 \\
             A-2DKF & First-order & SDP & 69 & 63.5 & 47.19 & 18.07 & 1205 \\
             A-2DKF & \textbf{High-order} & SDP & 71.7 & 58.4 & 47.62 & 8.24 & 651 \\
             \textbf{A-3DKF} & First-order & SDP & 71.9 & 65.1 & 49.6 & 13 & 1075 \\
             \textbf{A-3DKF} & \textbf{High-order} & SDP & \textbf{72.4} & \textbf{68.3} & \textbf{51.5} & \textbf{13.4} & \textbf{524} \\
        \end{tabular}
    }
\end{table*}
\paragraph{\textbf{Improvements with Active 3D Kalman Filter and High-order Association}} Firstly, we target on the use of object's 3D locations from our SODE in the motion model, i.e., the use of 3D information and the use of the acceleration in Kalman Filter as Active Kalman Filter; and the use of high-order association approach, i.e., considering the second-order relationship between objects within a frame in addition to the first-order similarity.

Table \ref{tab:ablation_KF_Association} shows the evaluation results on MOT-16 and MOT-17 validation set using provided bounding box annotations. 
By using either the proposed A-3DKF motion model or the high-order association approach, we can achieve about 2-3\% improvements on all MOT metrics. 
With the combination of both, the improvement is more than 3\% especially higher improvements for MT and ML which shows that it keeps tracking the pedestrian well with the use of two proposed modules, i.e., A-3DKF and high-order association.

 \paragraph{\textbf{Hyper-parameters in Cost Matrix}}
In this section, we conduct experiments on the impact of the choice of $\alpha$ and $\beta$ in Eq. \eqref{eq:affinity_matrix}. Table \ref{tab:ablation_hyperparameter_cost_matrix} shows that slightly more weight on first-order cost matrix than second-order cost matrix gives the best overall results than other configuration for $\alpha$ and $\beta$. Note that $\alpha = 1.0$ and $\beta = 0.0$ is the same as just only using first-order in the cost matrix for association.

\begin{table*}[!t]
    \footnotesize
    \caption{Evaluation on MOT-17 (SDP detectors) validation set with various choice of hyperparameters in computing affinitiy matrix. $\downarrow$ means the smaller the better; $\uparrow$ means the larger the better.}
    \label{tab:ablation_hyperparameter_cost_matrix}
    \centering
    \resizebox{\textwidth}{!}{
        \begin{tabular}{ll|cc|ccc}
            \textbf{ $\alpha$ } & \textbf{ $\beta$} & \textbf{MOTA} $\uparrow$ & \textbf{IDF1} $\uparrow$ & \textbf{MT} $\uparrow$ & \textbf{ML} $\downarrow$ & \textbf{ID Sw.} $\downarrow$ \\
            \hline
            \hline
            0.1 & 0.9 & 65.7 & 61.7 & 38.58 & 17.09 & 3137 \\
            0.3 & 0.7 & 66.0 & 61.7 & 41.36 & 16.48 & 3008 \\
            0.5 & 0.5 & 72.2 & 60.2 & 47.7 & \textbf{11.5} & 840 \\
            \textbf{0.6} & \textbf{0.4} & \textbf{72.4} & \textbf{68.3} & \textbf{51.5} & 13.4 & \textbf{524} \\
            0.7 & 0.3 & 71.2 & 59.5 & 42.3 & 15.3 & 861 \\
            0.9 & 0.1 & 72.0 & 51.9 & 44.6 & 17.3 & 875 \\
            1.0 & 0.0 & 71.9 & 65.1 & 49.6 & 13 & 1075 \\
        \end{tabular}
    }
\end{table*}

\begin{figure}
        \centering
        \includegraphics[width=0.8\columnwidth]{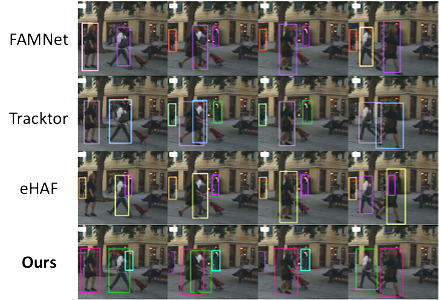}
        \caption{Tracking results in comparison with other methods (Tracktor \cite{bergmann2019tracking}, FAMNet \cite{chu2019famnet} and eHAF \cite{sheng2018heterogeneous})  on MOT challenge video MOT16-08. \textbf{Best viewed in color.}}
        \label{fig:example_tracking1}
\end{figure}

\subsection{Benchmark Evaluation}
\paragraph{\textbf{Comparisons with the state-of-the-art MOT systems on MOTChallenge}} We achieve the state-of-the-art results on both MOT16 and MOT17 testing set benchmarks using public detection and online settings. In this experiment, we adopt our best model setting with SODE and A-3DKF for motion model and high-order association. Table \ref{tab:MOT_results} shows the achieved results on these benchmark sets of MOT challenge. These results further emphasize the advantages of our DP-MOT with significant MOTA score improvements of 8.7\% and 9.3\% on MOT17 and MOT16, respectively. 
Qualitative results on several videos of MOT challenges are illustrated in Figs. \ref{fig:example_tracking1} and \ref{fig:example_tracking2}.
As one can see, our proposed DP-MOT can effectively handle the occlusions during the tracking process. Using the depth orders from SODE helps the later steps, i.e. high-order association computation, correctly associate the detections with track ids. For example, in Fig. \ref{fig:example_tracking1}, when two persons walk across each other, having the depth order of objects helps to avoid boxes switching from happening as in other approaches.

\begin{table*}[!t]
    \small
    \centering
    \caption{
    Comparisons among MOT methods with public detections on MOTChallenge test sets. $\downarrow$ means the smaller the better; $\uparrow$ means the larger the better.}
    \label{tab:MOT_results}
    \resizebox{\textwidth}{!}{
        \begin{tabular}{c|l|cc|cc}
             \textbf{Set} & \textbf{Method} & \textbf{MOTA} $\uparrow$ & \textbf{IDF1} $\uparrow$ & \textbf{MT} $\uparrow$ & \textbf{ML} $\downarrow$ \\
             \hline
            \hline
             \multirow{9}{*}{MOT17} & \textbf{DP-MOT (ours)} & \textbf{65.0} & \textbf{67.1} & 28.8 & \textbf{27.5} \\
             \multirow{2}{*}{}   & TMOH \cite{Stadler_2021_CVPR} & 62.1 & 62.8 & 26.9 & 31.4   \\
             \multirow{2}{*}{}   & Lif\_T \cite{lifted_disjoint_paths_2020_ICML} & 60.5 & 65.6 & 27.0 & 33.6   \\
             \multirow{2}{*}{}   & LPC\_MOT \cite{dai2021LPC} & 59.0 & 66.8 & \textbf{29.9} & 33.9   \\
             \multirow{2}{*}{}   & MPNTrack \cite{Braso_2020_CVPR} & 58.8 & 61.7 & 28.8 & 33.5   \\
              \multirow{2}{*}{}   & Tracktor++V2 \cite{bergmann2019tracking} & 56.3 & 55.1 & 21.1 & 35.3   \\
              \multirow{2}{*}{}   & FAMNet \cite{chu2019famnet} & 52.0 & 48.7 & 19.1 & 33.4   \\
              & eHAF \cite{sheng2018heterogeneous} & 51.8 & 54.7 & 23.4 & 37.9  \\
              & jCC \cite{keuper2018motion} & 51.2 & 54.5   & 20.9 & 37.0  \\
    
             \hline
             \multirow{9}{*}{MOT16} & \textbf{DP-MOT (ours)} & \textbf{63.7} & \textbf{66.2} & 25.8 & \textbf{28.2}  \\
             \multirow{2}{*}{}   & TMOH \cite{Stadler_2021_CVPR} & 63.2 & 63.5 & 27.0 & 31.0   \\
             \multirow{2}{*}{}   & Lif\_T \cite{lifted_disjoint_paths_2020_ICML} & 61.3 & 64.7 & 27.0 & 34.0   \\
             \multirow{2}{*}{}   & LPC\_MOT \cite{dai2021LPC} & 58.8 & 67.6 & \textbf{27.3} & 35.0   \\
             \multirow{2}{*}{}   & MPNTrack \cite{Braso_2020_CVPR} & 58.6 & 61.7 & \textbf{27.3} & 34.0   \\
              & Tracktor++ \cite{bergmann2019tracking} & 54.4 & 52.5 & 19.0 & 36.9   \\
              & KCF \cite{chu2019online} & 48.8 & 47.2 & 15.8 & 38.1  \\
              & GCRA \cite{ma2018trajectory} & 48.2 & 48.6 & 12.9 & 41.1  \\
              & MOTDT \cite{chen2018real} & 47.6 & 50.9 & 15.2 & 38.3  \\ 
        \end{tabular}
    }
\end{table*}

\paragraph{\textbf{Comparisons with the state-of-the-art MOT systems on nuScene dataset}}
Our goal in this experiment is to demonstrate the generalization of our proposed method. It is not only applicable to static camera settings but also in dynamic camera settings, i.e. autonomous driving scenarios, on nuScene dataset \cite{caesar2020nuscenes}. We also compare our approach with other vision-based MOT approaches, including CetnerTrack \cite{zhou2020tracking} and DEFT \cite{Chaabane2021deft} 
on nuScene dataset. Our proposed approach achieves competitive results with other MOT approaches. 

\begin{table*}[!t]
    \small
    \centering
    \caption{Comparison of MOT performance on the nuScenes \textbf{validation set} for Vision Track challenge.}
    \resizebox{1.0\textwidth}{!}{
    \begin{tabular}{|l|c|c|c|c|c|c|c|c|c|c|c|}
    \hline
       \textbf{Method} & \textbf{AMOTA}	& \textbf{AMOTP}	& \textbf{MOTAR} &	\textbf{MOTA} $\uparrow$ & \textbf{MOTP} $\uparrow$ & \textbf{RECALL} $\uparrow$ &	\textbf{MT} $\uparrow$ &	\textbf{ML} $\downarrow$& \textbf{IDS} $\downarrow$	& \textbf{FRAG} $\downarrow$ \\	
       \hline       
       CenterTrack \cite{zhou2020tracking} & 0.068 & 1.543 & 0.349 & 0.061 & 0.778 & 0.222 & 524 & 4,378 & 2,673 & 1,882 \\ 
       DEFT \cite{Chaabane2021deft} & 0.213 & 1.532 & 0.49 & 0.183 & 0.805 & 0.4 & 1,591 & 2,552 & 5,560 & 2,721 \\ 
       \hline
       \textbf{Ours} & \textbf{0.242} & 1.541 & 0.568 & 0.197           & \textbf{0.832}             & \textbf{0.453}             & \textbf{1,643}          & \textbf{2,162}            & \textbf{1,362}             & \textbf{1,462} \\
      \hline
    \end{tabular}
    }
    \label{tab:nuscene_val_track_results}
\end{table*}

\section{Conclusions}

This work has presented a simple yet efficient subject-ordered depth estimation to tackle the occlusion problem in MOT by order the depth positions of detected subjects.
Then the new Active pseudo-3D Kalman filter was proposed to keep updating the movement and order of objects dynamically. 
In addition, we have shown that introducing first-order and second-order relationships between the tracks and detected objects to the data association steps helps to boost the performance to consistently achieve the state-of-the-art results on the MOT challenges. 
We hope that our simple SODE will foster future work in studying what kind of prior knowledge is available in off-the-shelf detectors. Due to the limitation of optical flow, people tracking in a crowd scene can be slower than a normal scene.
Our future work aims to study the impact of various 2D detectors on the accuracy of depth orders and how to extend and improve recent end-to-end 2D-MOT approaches using a similar pseudo-3D idea.
\newpage

\appendix

\section{Method Details}

In this section, we provide more details on the proposed methods, i.e. A-3DKF and SODE, and network structure for feature embedding.

\subsection{Feature Embedding Network}

\paragraph{Network Structure}

The Feature Embedding Network (FEN), a deep convolutional neural network, is used to extract appearance features $\mathbf{e}$ for each detected box. 
FEN is based on the network employed in \cite{wang2019towards} where we only use its feature embedding branch, and modify its inputs to take detected boxes and provided features of the corresponding boxes. The backbone of FEN is DarkNet-53 \cite{redmon2018yolov3} to provide a balance between accuracy and speed. Feature Pyramid Network \cite{lin2017feature} structure is applied to the backbone to handle objects at different scales. 
The feature embedding branch is connected with the backbone structure by a convolution layer with 512 channels to obtain identity embedding features at each location and scale. The embedding feature vector $\mathbf{e}$ $\in \bbR^{512 \times 1}$ is extracted from the resulting feature map.

\paragraph{Comparison of Network Structure for Feature Embedding} In this section, we compare the choice of deep network structures, i.e. in terms of input sizes, in the feature sub-net (FEN). Each detected box will be cropped from an image and resized to fit the input of the network. We perform this experiment using three different input sizes, i.e. small ($32 \times 96$), medium ($45 \times 135$), and large ($64 \times 192$). Table \ref{tab:network_input_size} shows the results on MOT-16 training set to demonstrate the effects of the input sizes on the MOT results. 
The higher resolutions mean better appearance features being extracted but the network is more complex. To balance between the accuracy and the speed, we can choose an appropriate input size accordingly. The running speed of each configuration is 46 fps, 38 fps, and 30 fps for FEN-Small, FEN-Medium, and FEN-Large, respectively. For the best performance, we use the large input size in all other experiments.

\begin{table}[!h]
    \centering
    \resizebox{\textwidth}{!}{
    \begin{tabular}{l|cc|ccc|c}
     \begin{tabular}{@{}c@{}}\textbf{Network} \\ \textbf{Structure}\end{tabular}
    & \textbf{MOTA} $\uparrow$ & \textbf{IDF1} $\uparrow$ & \textbf{MT} $\uparrow$ & \textbf{ML} $\downarrow$ & \textbf{ID Sw.} $\downarrow$ & Speed (fps) $\uparrow$ \\
     \hline
     FEN-Large & \textbf{55.4} & \textbf{54.7} & \textbf{27} & \textbf{26} & \textbf{789} & 30 \\
     FEN-Medium & 54 & 54 & 24.5 & 29.3 & 806 & 38 \\
     FEN-Small & 52.1 & 55.0 & 23.6 & 31.5 & 1116 & \textbf{46} \\
    \end{tabular}
    }
    \caption{Various configurations of Feature Embedding sub-net evaluated on MOT-16 training set. $\downarrow$ means the smaller the better; $\uparrow$ means the larger the better.}
    \label{tab:network_input_size}
\end{table}

\subsection{Occlusion Handling with A-3DKF}
\begin{figure} [!t]
    \centering
    \includegraphics[width=1.0\columnwidth]{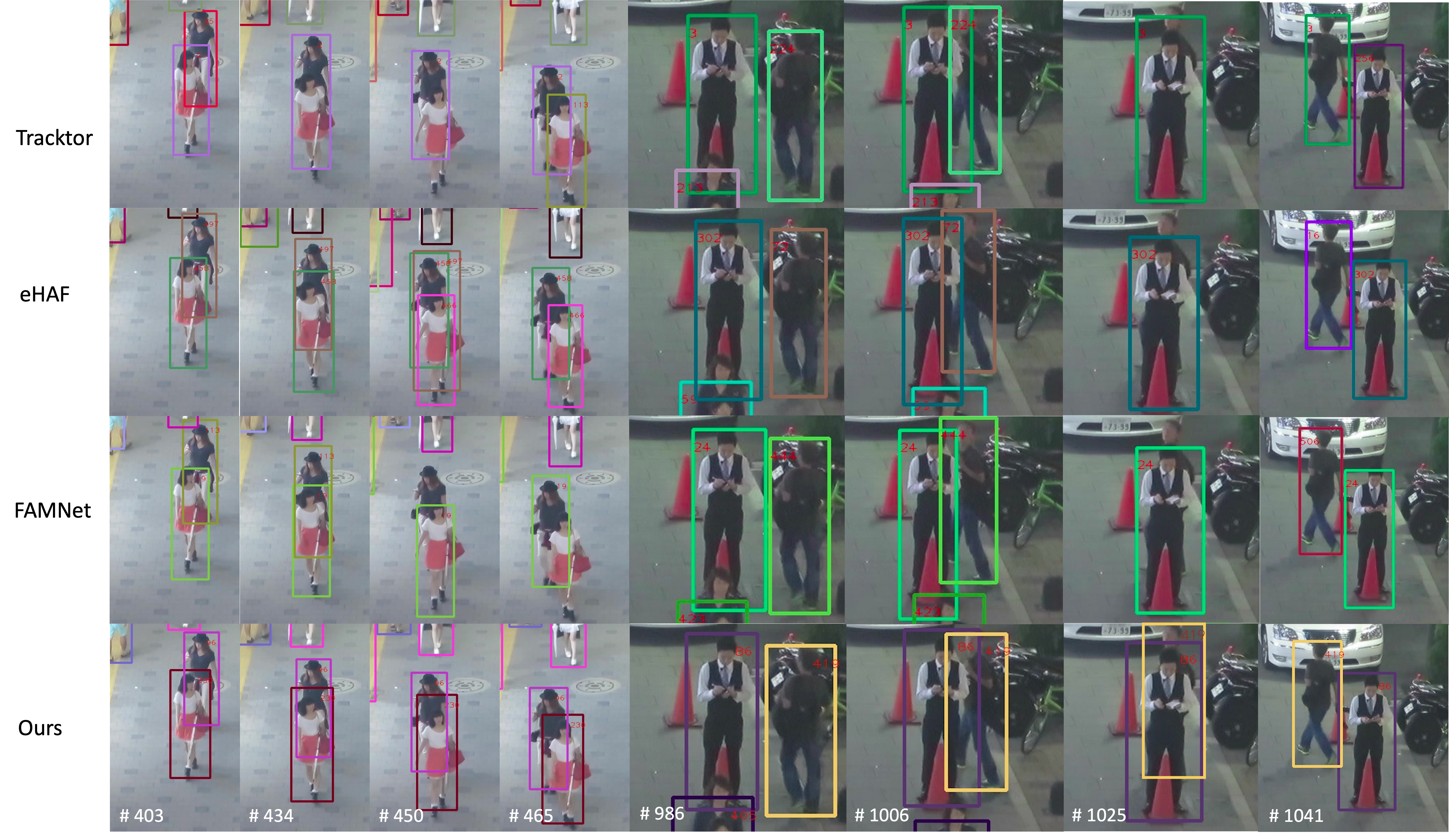}
    \caption{An example of how our proposed approach (the last row) successfully handles a large overlapping case in MOT Challenge Video comparing with the state-of-the-art approaches Tracktor \cite{bergmann2019tracking}, eHAF \cite{sheng2018heterogeneous}, and FAMNet \cite{chu2019famnet} (other rows). \textbf{Best viewed in color.}}   
     \label{fig:occlusion_example}
\end{figure}

One observation from our experiments is that most methods fail due to the detector mistakenly identifies two targeted subjects as one, i.e. a bounding box cover both two objects or each object partially when a large overlapping occurs. (See Fig. \ref{fig:occlusion_example}).
Such cases can be detected by checking the column of matrix $\bC$ containing the two smallest values where the difference is smaller than threshold $\tau_C$.
We will not use the detected box to update the state of A-3DKF, we will keep the two corresponding tracks activated.

\section{Implementation Details}

\paragraph{Parameters}

The weights combining first-order and frame-wise cost matrix, i.e. $\alpha$ and $\beta$, are experimentally set to $0.45$ and $0.55$, respectively. The threshold for distance in depth $\tau_z$ is set to $\frac{1}{10} \times H $, where $H$ is the height of the input image.
 
\paragraph{Dataset splits}

FEN is trained on multiple large-scale datasets with pedestrian identity annotations which were described in \cite{wang2019towards}. These pedestrian datasets, which contain both bounding box and identity annotations, are Caltech \cite{dollar2009pedestrian}, MOT-16 training set \cite{MOT16}, CUHK-SYSU \cite{xiao2017joint}, and PRW \cite{zheng2017person}.

For evaluation tracking performance on pedestrians, we use a dataset published by MOT challenges including MOT 2016 and MOT 2017. 
We use the validation set of MOT16 (including \textit{MOT16-02, MOT16-04, MOT16-05, MOT16-09, MOT16-10, MOT16-11 and MOT16-13} videos), for ablation study and parameter configuration.
The testing set of MOT16 (including \textit{MOT16-01, MOT16-03, MOT16-06, MOT16-07, MOT16-08, MOT16-12 and MOT16-14} videos)  and MOT17 (including \textit{MOT17-01, MOT17-03, MOT17-06, MOT17-07, MOT17-08, MOT17-12 and MOT17-14} videos) are used to generate results and submit to challenge for Tab. \ref{tab:MOT_results}.

\section{Additional Results}

\begin{figure}[!h]
    \centering
    \includegraphics[width=1\columnwidth]{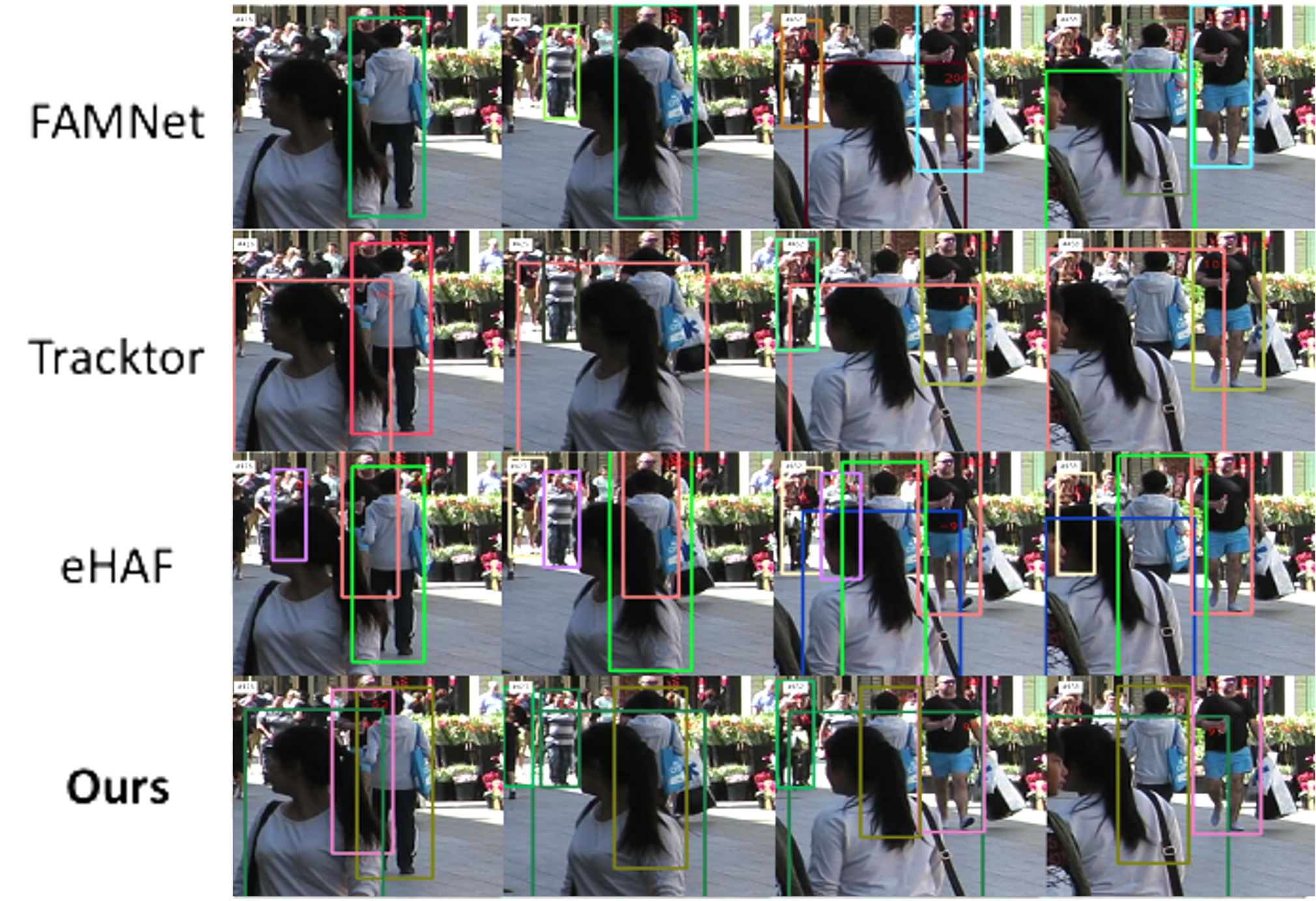}
    \caption{Tracking results in comparison with other methods (Tracktor \cite{bergmann2019tracking}, FAMNet \cite{chu2019famnet} and eHAF \cite{sheng2018heterogeneous})  on MOT challenge video MOT16-08. \textbf{Best viewed in color.}}
    \label{fig:example_tracking2}
\end{figure}

In this section, we illustrate more results compared with other SOTA methods in Fig. \ref{fig:example_tracking2}. We also show the results of individual sequences in the MOT-17 challenge in Table \ref{tab:detail_mot_17}.

\begin{table*}[!t]
    \small
    \centering
    \caption{
    Results of individual sequences in MOTChallenge MOT-17 test sets. $\downarrow$ means the smaller the better; $\uparrow$ means the larger the better.}
    \label{tab:detail_mot_17}
    \resizebox{\textwidth}{!}{
        \begin{tabular}{c|l|cc|cc}
             \textbf{Sequences} & \textbf{Camera Setting} & \textbf{MOTA} $\uparrow$ & \textbf{IDF1} $\uparrow$ & \textbf{MT} $\uparrow$ & \textbf{ML} $\downarrow$ \\
            \hline
            \hline
          MOT17-01 & static & 54.9 & 63.6 & 7 & 5 \\
          MOT17-03 & static & 81.9 & 78.6 & 106 & 11 \\
          MOT17-08 & static & 36.3 & 40.7 & 16 & 30 \\
          \hline
          MOT17-06 & handheld & 59.8 & 60.0 & 69 & 59 \\
          MOT17-07 & handheld & 59.4 & 57.8 & 15 & 14 \\
          MOT17-12 & handheld & 53.8 & 59.3 & 20 & 35 \\
          \hline
          MOT17-14 & vehicle & 45.3 & 55.2 & 17 & 51 \\
        \end{tabular}
    }
\end{table*}

\newpage

\bibliographystyle{model2-names}
\bibliography{refs}

\clearpage
\clearpage

\section*{Author Biographies}

\begin{wrapfigure}{L}{0.3\textwidth}
\centering
\includegraphics[width=0.25\textwidth]{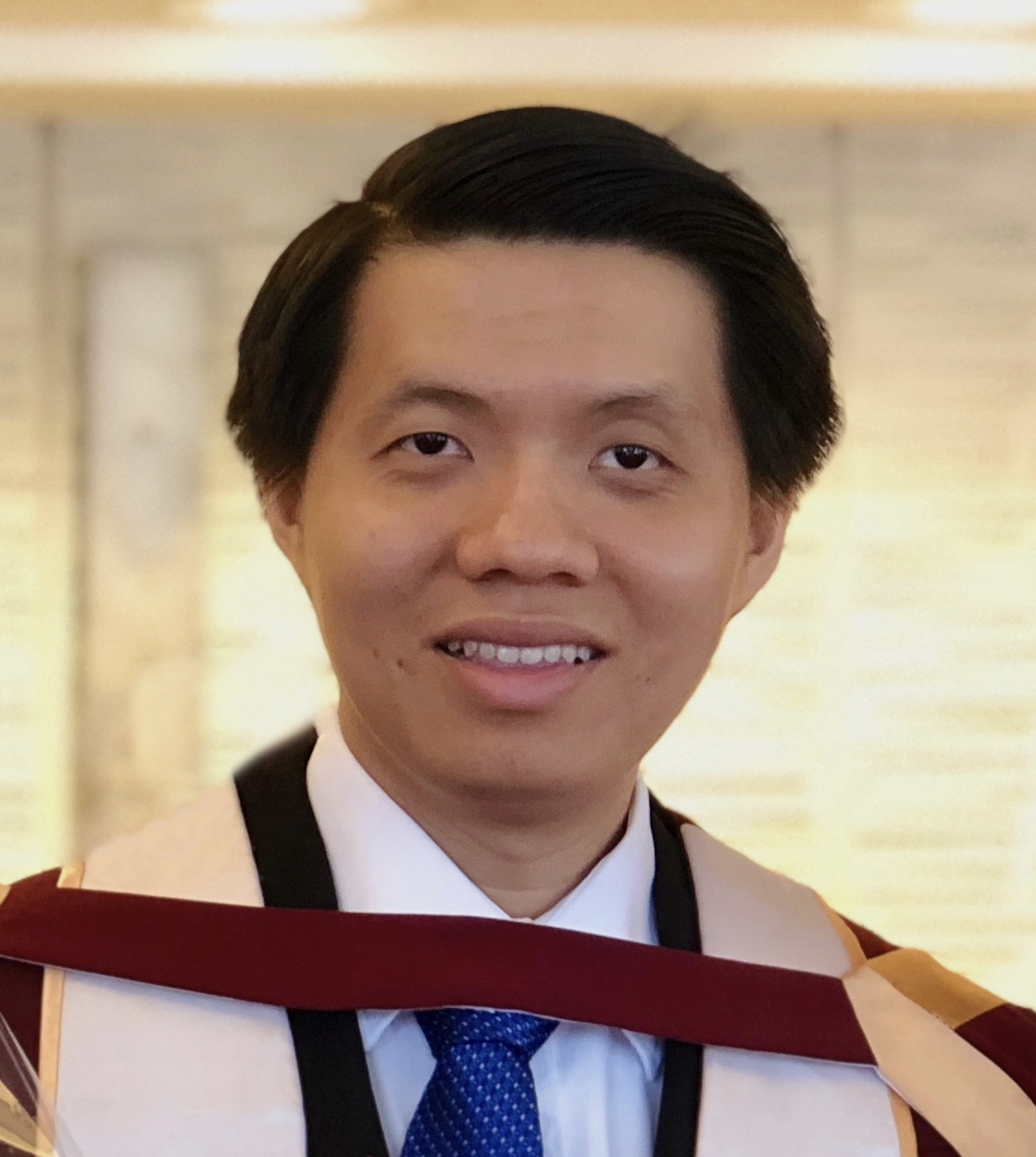}
\end{wrapfigure}
\noindent {\bf Kha Gia Quach} is currently a Senior Technical Staff while having research collaboration with both Computer Vision and Image Understanding (CVIU) Lab, University of Arkansas, USA, and Concordia University, Montreal, Canada. He has been a research associate in Cylab Biometrics Center at Carnegie Mellon University (CMU), USA since 2016. He received his Ph.D. degree in Computer Science under supervision of Prof. Tien Dai Bui and Prof. Khoa Luu from the Department of Computer Science and Software Engineering, Concordia University, Montreal, Canada, in 2018. He received his B.S. and M.Sc. degree in Computer Science from the University of Science, Ho Chi Minh City, Vietnam, in 2008 and 2012, respectively. His research interests lie primarily in Compressed Sensing, Sparse Representation, Image Processing, Machine Learning, and Computer Vision.
He has published and co-authored over 20 papers in top conferences including CVPR, BTAS, ICPR, ICASSP, and premier journals including TIP, PR, CVIU, IJCV, CJRS. 
He has been a reviewer of several top-tier journals and conferences including TPAMI, TIP, SP, CVPR, ICCV, ECCV, AAAI, MICCAI. He also served as a Program Committee Member of Precognition: Seeing through the Future, CVPR 2019, and 2020.

\begin{wrapfigure}{L}{0.3\textwidth}
\centering
\includegraphics[width=0.25\textwidth]{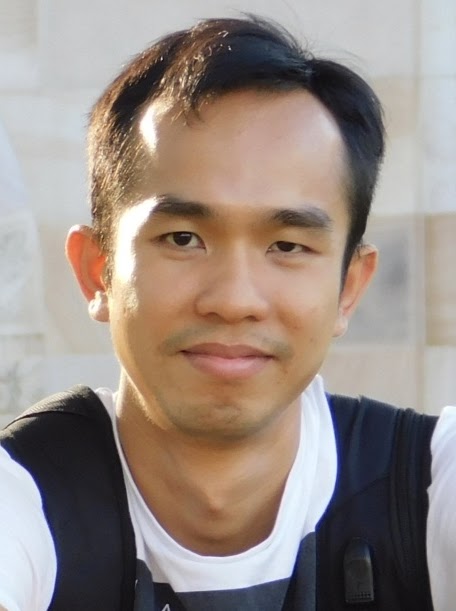}
\end{wrapfigure}
\noindent {\bf Huu Le} received the BS degree in electrical engineering from Portland State University, Oregon, in 2011, and the PhD degree in computer science from the University of Adelaide, in 2018. He is currently a postdoctoral researcher with the Electrical Engineering Department, Chalmers University of Technology, Sweden. His research interests include robust estimation, computational geometry, and numerical optimization methods applied to the fields of computer vision.

\clearpage
\begin{wrapfigure}{L}{0.3\textwidth}
\centering
\includegraphics[width=0.25\textwidth]{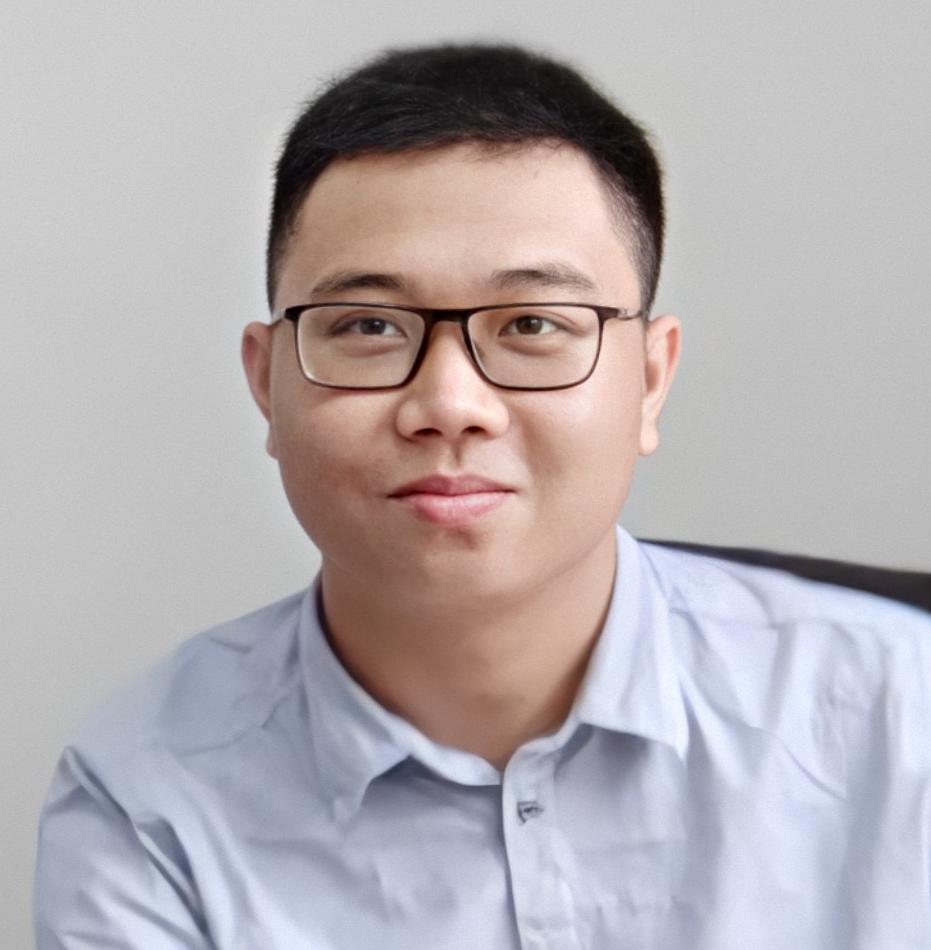}
\end{wrapfigure}
\noindent {\bf Pha Nguyen} is currently a Ph.D student in Computer Science and a Research Assistant at Computer Vision and Image Understanding (CVIU) Lab, University of Arkansas, USA. In 2020, he obtained his B.Sc degree from Vietnam National University Ho Chi Minh City - University of Science in Computer Science. His particular research interests are focused on deep learning, machine learning, and their applications in computer vision, primarily multiple object tracking. He also serves as a reviewer for ECCV and IEEE Access. \\

\begin{wrapfigure}{L}{0.3\textwidth}
\centering
\includegraphics[width=0.25\textwidth]{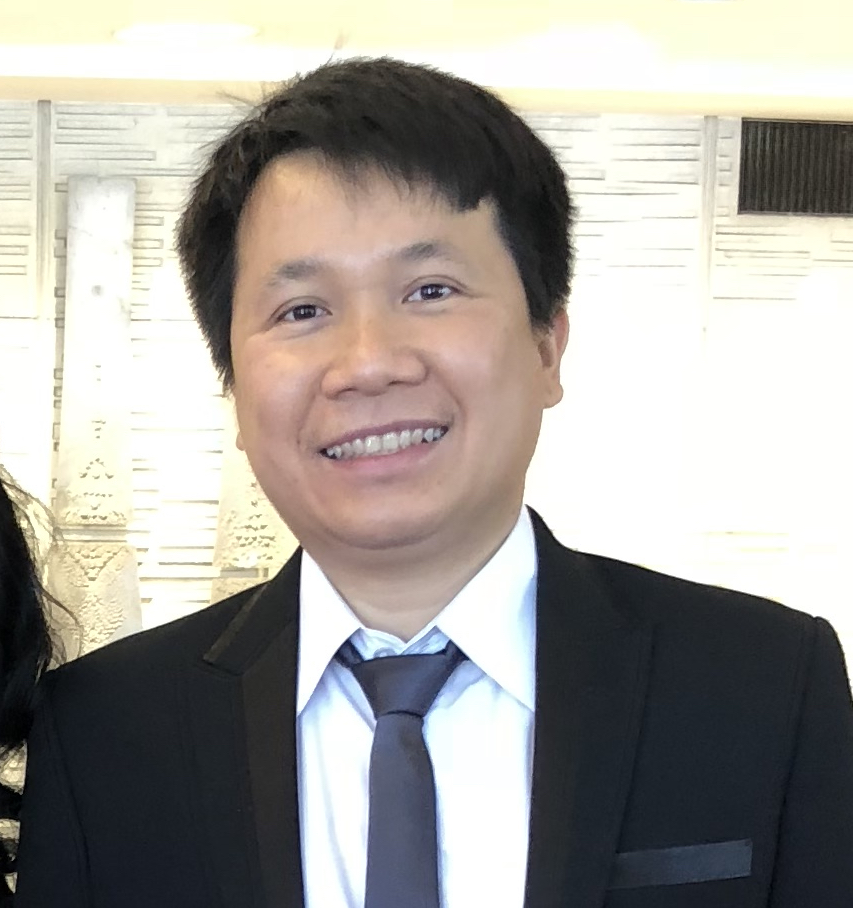}
\end{wrapfigure}
\noindent \textbf{Chi Nhan Duong} is currently a Senior Technical Staff and having research collaborations with both Computer Vision and Image Understanding (CVIU) Lab, University of Arkansas, USA, and Concordia University, Montreal, Canada.
He had been a Research Associate in Cylab Biometrics Center at Carnegie Mellon University (CMU), USA since September 2016. He received his Ph.D. degree in Computer Science under the supervision of Prof. Tien Dai Bui and Prof. Khoa Luu from the Department of Computer Science and Software Engineering, Concordia University, Montreal, Canada. He was an Intern with National Institute of Informatics, Tokyo Japan in 2012. He received his B.S. and M.Sc. degrees in Computer Science from the Department of Computer Science, Faculty of Information Technology, University of Science, Ho Chi Minh City, Vietnam, in 2008 and 2012, respectively.    
His research interests include Deep Generative Models, Face Recognition in surveillance environments, Face Aging in images and videos, Biometrics, and Digital Image Processing, and Digital Image Processing (denoising, inpainting, and super-resolution).
He is currently a reviewer of several top-tier journals  including TPAMI, TIP, SP, PR, PR Letters. He is also recognized as an outstanding reviewer of several top-tier conferences such as CVPR, ICCV, ECCV, AAAI. He is also a Program Committee Member of Precognition: Seeing through the Future, CVPR 2019 and 2020. \\

\begin{wrapfigure}{L}{0.3\textwidth}
\centering
\includegraphics[width=0.25\textwidth]{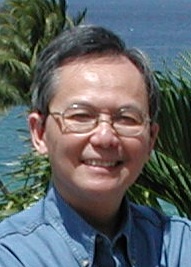}
\end{wrapfigure}
\noindent \textbf{Tien D. Bui} He is a full professor in the Department of Computer Science and Software Engineering at Concordia University, Montreal, Canada. He was Chair of the Department from 1985 to 1990 and Associate Vice-Rector Academic (Research) from 1992 to 1996. He  served on various governing bodies of Concordia University including its Senate and the Board of Governors as well as the Boards of Directors of many research centers and institutes in Quebec. He is Associate Editor of Signal Processing (EURASIP) and International Journal of Wavelets, Multiresolution and Information Processing. He has
served on many organizing and program committees of international conferences as well as a member of many grant selection committees at both provincial and federal levels. He has published more than 200 papers in different areas and is a co-author of the book Computer Transformation of Digital Images and Patterns (World Scientific 1989). His current interests include machine learning, pattern recognition, image, and video processing. \\

\begin{wrapfigure}{L}{0.3\textwidth}
\centering
\includegraphics[width=0.25\textwidth]{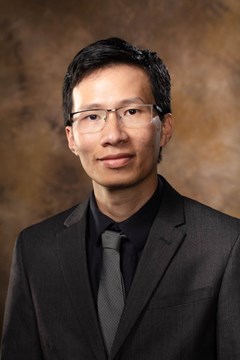}
\end{wrapfigure}
\noindent \textbf{Khoa Luu} Dr. Luu is currently an Assistant Professor and the Director of Computer Vision and Image Understanding (CVIU) Lab in Department of Computer Science \& Computer Engineering at University of Arkansas, Fayetteville. He is serving as an Associate Editor of IEEE Access journal. He was the Research Project Director in Cylab Biometrics Center at Carnegie Mellon University (CMU), USA. He has received four patents and two best paper awards and coauthored 100+ papers in conferences and journals. He was a vice-chair of Montreal Chapter IEEE SMCS in Canada from September 2009 to March 2011.
He is teaching Computer Vision, Image Processing, and Introduction to Artificial Intelligence courses in CSCE Department at the University of Arkansas, Fayetteville.
His research interests focus on various topics, including Biometrics, Image Processing, Computer Vision, Machine Learning, Deep Learning, Multifactor Analysis, and Compressed Sensing. 
He is a co-organizer and a chair of the CVPR Precognition Workshop in 2019, 2020, 2021; MICCAI Workshop in 2019, 2020 and ICCV Workshop in 2021. He is a PC member of AAAI, ICPRAI in 2020 and 2021.
He is currently a reviewer for several conferences and journals, such as CVPR, ICCV, ECCV, NeurIPS, ICLR, FG, BTAS, IEEE-TPAMI, IEEE-TIP, Journal of Pattern Recognition, Journal of Image and Vision Computing, Journal of Signal Processing, Journal of Intelligence Review, IEEE Access Trans., etc.

\end{document}